\documentclass[12pt]{article}
\usepackage{amsmath}
\usepackage{graphicx}
\usepackage{enumerate}
\usepackage{natbib}
\usepackage[hyphens]{url} 
\usepackage{algorithm} 
\usepackage{algpseudocode} 
\usepackage{colortbl}
\usepackage{inconsolata}
\usepackage{dsfont}
\usepackage{adjustbox}
\usepackage{amsmath}
\usepackage{amssymb}
\usepackage{csquotes}\MakeOuterQuote{"}
\usepackage{hyperref}
\usepackage{xcolor}
\definecolor{darkblue}{rgb}{0, 0, 0.5}
\hypersetup{colorlinks=true,citecolor=darkblue, linkcolor=darkblue, urlcolor=darkblue}
\usepackage{longtable}

\newcommand{\blind}{0}

\usepackage{lineno}

\begin{document}

\def\spacingset#1{\renewcommand{\baselinestretch}%
{#1}\small\normalsize} \spacingset{1}
\setlength{\footnotesep}{2\footnotesep}


\if0\blind
{
  \title{\bf Estimating the Influence of Sequentially Correlated Literary Properties in Textual Classification: A Data-Centric Hypothesis-Testing Approach}
  \author{Gideon Yoffe\\
    Department of Statistics and Data Science, Hebrew University of Jerusalem\\
    \and 
    Nachum Dershowitz \\
    School of Computer Science and AI, Tel Aviv University \\
   \and  
    Ariel Vishne \\
    Department of Statistics and Data Science, Hebrew University of Jerusalem \\
    \and
    Barak Sober \\
    Department of Statistics and Data Science, Hebrew University of Jerusalem
    }
  \maketitle
} \fi

\if1\blind
{
  \bigskip
  \bigskip
  \bigskip
  \begin{center}
    {\LARGE\bf Estimating the Influence of Sequentially Correlated Literary Properties in Textual Classification: A Data-Centric Hypothesis-Testing Approach}
\end{center}
  \medskip
} \fi

\bigskip

\begin{abstract}
We introduce a data-centric hypothesis-testing framework to quantify the influence of sequentially correlated literary properties—such as thematic continuity—on textual classification tasks. Our method models label sequences as stochastic processes and uses an empirical autocovariance matrix to generate surrogate labelings that preserve sequential dependencies. This enables statistical testing to determine whether classification outcomes are primarily driven by thematic structure or by non-sequential features like authorial style.

Applying this framework across a diverse corpus of English prose, we compare traditional (word $n$-grams and character $k$-mers) and neural (contrastively trained) embeddings in both supervised and unsupervised classification settings. Crucially, our method identifies when classifications are confounded by sequentially correlated similarity, revealing that supervised and neural models are more prone to false positives -- mistaking shared themes and cross-genre differences for stylistic signals. In contrast, unsupervised models using traditional features often yield high true positive rates with minimal false positives, especially in genre-consistent settings.

By disentangling sequential from non-sequential influences, our approach provides a principled way to assess and interpret classification reliability. This is particularly impactful for authorship attribution, forensic linguistics, and the analysis of redacted or composite texts, where conventional methods may conflate theme with style. Our results demonstrate that controlling for sequential correlation is essential for reducing false positives and ensuring that classification outcomes reflect genuine stylistic distinctions.

\end{abstract}

\noindent%
\vfill

\newpage
\spacingset{1.9} 
\section{Introduction}
\label{sec:intro}

When considering the process of text composition, numerous properties come into play, with genre, style, and theme being particularly significant. Identifying the primary factor distinguishing texts among these three is crucial for tasks like authorship attribution and forensic linguistics \citep{holmes1994authorship, koppel2004authorship}.

Style, theme, and genre are fundamental properties that play distinct yet intertwined roles in the composition and interpretation of textual content. Style refers to the distinctive manner in which a writer expresses ideas, characterized by choices in vocabulary, sentence structure, tone, and rhetorical devices. It encompasses the unique linguistic fingerprint of an author and contributes to the overall aesthetic and readability of the text \citep[e.g.,][]{saukkonen2003define}. The theme, conversely, pertains to the central ideas, messages, or subjects explored within the text. It reflects the underlying concepts or motifs that recur throughout the narrative, conveying deeper meaning and resonance to the reader \citep[e.g.,][]{brinker1995theme}. Genre, meanwhile, categorizes texts into distinct literary or textual forms based on shared conventions, structures, and expectations \citep[e.g.,][]{chandler1997introduction}. It provides a framework for understanding and classifying texts according to their overarching narrative structures, plot elements, and stylistic conventions.

In computational linguistics, characterizing specific desired properties of a text -- such as theme, style, or genre -- poses significant challenges due to their abstract and multifaceted nature \citep{rybicki2016computational}. These parameters encompass a broad range of linguistic elements, making them difficult to quantify and model computationally. Furthermore, these parameters vary widely across authors and texts, posing challenges for developing universal computational models \citep{serrano2009modeling}. Identifying patterns and connections within these parameters involves deciphering complex semantic and conceptual relationships, which can be context-dependent and subject to individual interpretation. This variability makes it challenging to automate analysis reliably across diverse texts.
Additionally, the classification of texts based on these parameters faces similar challenges, as texts often defy traditional categorization and exhibit hybrid or evolving conventions. Additionally, such classification relies on intricate intertextual and extratextual factors, such as cultural context and reader expectations, which are challenging to formalize computationally. Overall, literary works' abstract and multidimensional nature presents formidable obstacles in computational linguistics, demanding innovative approaches and interdisciplinary collaborations for adequate characterization.

Of particular interest is the discipline of stylometry, which presupposes the possibility of distinguishing between distinct authorial styles (that are independent of thematic content) based on statistical learning analyses \citep{kestemont2014function}. 
Identifying such a stylistic property is helpful in a variety of real-world settings (e.g., forensic linguistics \citep{LambersForensicAA}, plagiarism detection \citep{ainsworth2018wrote}, and authorship analysis \citep{Kestemont2016} -- where it has been shown that other textual properties are often misleading in asserting authorship \citep{Stamatatos2009_review, stamatatos2018masking}). 

Stylometric approaches are typically categorized into supervised and unsupervised settings \citep{juola2008authorship}. In supervised tasks, a set of authors and their works are provided for training, with the goal of attributing an unattributed text to one of the authors—known as authorship attribution. Two assumptions can apply: (\textbf{1}) the author must be among the provided set \citep{Koppel2009_AA_AV_review}, or (\textbf{2}) the author may not be in the set \citep{Koppel2011_in_the_wild}. Alternatively, authorship verification aims to determine if an unknown work was written by a known author or someone else \citep[e.g.,][]{Koppel2007_unmasking, juola2008authorship, Kestemont2016}. Other supervised tasks include authorship profiling, where texts are attributed to a shared group profile (e.g., historical era, gender, language) \citep{Argamon2009}, as well as cross-genre and cross-domain settings \citep[e.g.,][]{Kestemont2012_cross_genre_AV, stamatatos2018masking}. These tasks have been applied across different languages, eras, and text sizes \citep{Schwartz2013_short_texts}.

In the unsupervised setting, no prior author information is given. The task is to identify whether pairs of unlabeled texts (or text units like sentences or paragraphs) were written by the same author. The core question is: \textit{Did the same author write these texts?}

This work introduces a novel approach to identifying whether sequentially correlated literary properties contribute to text classification. We apply a hypothesis-testing framework to analyze such correlations in label sequences, which are often linked to thematic content, rather than authorial style. Our stochastic model of sequential correlations offers an empirical basis for distinguishing between \textit{stylistic} versus certain \textit{non-stylistic} influences.

We test our approach on English prose texts of varying genre similarity, authored either by the same or by different authors. Using both unsupervised and supervised methods with traditional and state-of-the-art neural embeddings -- optimized for capturing authorial style, we apply our hypothesis-testing framework to assess whether the classification is primarily driven by sequential correlations, as seen in same-author, genre-similar texts, or by other factors.
These experiments affirm that accounting for the influence of sequential correlations in textual classification consistently improves the distinction between stylistic and non-stylistic classification outcomes in all the setups we tried.

This research is motivated by the challenges posed by multilayered and redacted historical texts, which require interdisciplinary approaches across linguistics, history, and computer science to trace their evolution from origin to extant form \citep[e.g.,][]{buhler2023exploring}. These texts frequently include annotations, marginalia, and modifications made by scribes or editors, obscuring the original authorial intent \citep[e.g.,][]{muller2014evidence}, and are often written in archaic or obsolete languages that complicate automated analysis \citep[e.g.,][]{baum2017content}. Additionally, their interpretation demands careful consideration of historical context, including cultural norms, linguistic conventions, and socio-political factors, to ensure accurate analysis \citep{chapman2017historical}. While our current work does not directly address historical texts, it seeks to validate our approach by applying it to modern texts with undisputed authorial and stylistic qualities, thereby providing a robust foundation for future applications to more complex historical material.


\section{The State of Stylometry} \label{previous_work}

Stylometry, the quantitative study of literary style, can be traced to the mid-20th century when scholars such as Mosteller and Wallace applied statistical methods to assess the authorship of the \textit{Federalist Papers}, demonstrating the effectiveness of function word frequencies and statistical inference in distinguishing authorship \citep{mosteller1963inference}. This pioneering work provided early evidence that consistent, measurable linguistic signals could be exploited to resolve long-standing literary and historical debates, laying the groundwork for contemporary computational stylometry.

With the growth of computational linguistics and data-driven approaches in the late 20th and early 21st centuries, the scope of stylometric research broadened considerably. Beyond basic authorship attribution, researchers developed models for authorship verification, author profiling, genre classification, and clustering of anonymous or disputed texts \citep[e.g.,][]{Juola2006, juola2008authorship, rybicki2013stylistics}. \textbf{One influential contribution from this period is Burrows’ Delta method \citep{burrows2002delta}, which quantifies stylistic distance using Z-score-based comparisons of function word frequencies. Its simplicity, language independence, and empirical robustness have made it a foundational benchmark in both literary and forensic stylometry.} As a widely adopted baseline for measuring authorial similarity, Delta offers a natural point of comparison for the present work, which introduces a sequential correlation-based framework aimed at disentangling stylistic signals from theme-driven textual structure.

A landmark machine learning-oriented approach by \citet{Koppel2007_unmasking} demonstrated the robustness of stylometric classification using frequent features such as common lexical items, function words, and character-level statistics. Their work laid the foundation for high-accuracy authorship identification across genres and languages. Further refinements—incorporating punctuation patterns \citep{darmon2021pull} and syntactic features—showed that such frequent signals could capture authorial style more reliably than topic- or content-driven lexical features, which are often more variable and less discriminative.

Nonetheless, growing evidence has shown that traditional frequency-based features can be confounded by thematic content. \citet{mikros2007investigating} demonstrated that many widely-used stylometric variables are topic-sensitive, and may mislead attribution systems when stylistic and thematic patterns overlap. This has led to a critical reassessment of assumptions underlying feature selection. Studies such as \citet{savoy2013feature}, \citet{lagutina2019survey}, and \citet{grieve2023register} argue that true stylistic modeling requires more than frequency analysis—it requires incorporating structural, syntactic, and discourse-level features capable of representing deeper authorial patterns.

To address these limitations, several works have proposed syntactic and structural approaches. For example, \citet{feng2012syntactic} employed syntactic parse tree structures to detect deception and authorial signals, while \citet{hollingsworth2012using} showed that dependency parsing can help distinguish authorial style from topic. These syntactic methods provide a level of abstraction that is less sensitive to thematic variation and more robust across genres and text types.

Information-theoretic and compression-based approaches have also gained traction, particularly due to their robustness and language-agnostic nature. \citet{cilibrasi2005clustering} introduced normalized compression distance (NCD), a compression-based measure of similarity that operates without predefined linguistic features or language-specific preprocessing. By quantifying the shared information content between two texts using standard compression algorithms, NCD provides a robust, domain-agnostic method for clustering and comparison. This approach offers a valuable alternative to traditional feature-engineered methods, and its emphasis on holistic textual similarity complements the data-centric perspective of the current work—particularly in highlighting classification signals that emerge from structural regularities rather than explicitly engineered feature sets.

Stylometry has also become increasingly relevant in forensic linguistics, where the need for interpretable, defensible results is paramount. \citet{juola2007future} emphasized the importance of transparency and reproducibility in authorship analysis, especially in legal or historical contexts where the evidentiary burden is high. More recent work has focused on evaluating the reliability and statistical significance of classification results, often calling for hypothesis-driven models that can quantify uncertainty and isolate confounding factors.

The rise of neural language models has brought a new wave of high-performance tools to stylometry. Studies such as \citet{ding2017learning} and \citet{canbay2020deep} explore deep learning models trained for stylometric tasks, including authorship attribution and genre classification. These models leverage high-dimensional contextual embeddings, often derived from large pre-trained transformers. While they show strong empirical results, they are frequently criticized for their lack of interpretability and for their potential sensitivity to thematic and structural artifacts \citep{dror2018hitchhiker, dror2020statistical}.

Despite these methodological advances, a major gap persists in understanding what type of information—style, theme, or genre—drives classification success. \citet{litvinova2020stylometrics}, \citet{hou2020robust}, and \citet{schuster2020limitations} highlight the risks of misattributing thematic similarity as stylistic consistency. This concern is especially salient in historical, redacted, or multi-author texts where thematic coherence may coexist with significant stylistic variation. Without tools to differentiate these axes, stylometric analyses risk drawing misleading conclusions.

This distinction is especially crucial in the analysis of complex or redacted texts, such as those found in the biblical corpus. Many of the books in the Hebrew Bible, for instance, are widely believed to be composite, incorporating contributions from multiple authorial sources, traditions, and editorial strata over time \citep[e.g.,][]{wellhausen1885prolegomena, holzinger1893einleitung, Gunkel1895}. Within the framework of the documentary hypothesis, texts such as the Book of Exodus and the Book of Genesis have been segmented into distinct sources (e.g., Priestly, Elohistic, Yahwistic) based on recurring stylistic, lexical, and thematic features. Yet, these divisions are often based on assumptions that thematic consistency necessarily reflects authorship or redaction \citep{wellhausen1885prolegomena, albertz2018recent}. Recent work has sought to bring more computational rigor to these debates. For example, \citet{dershowitz2015computerized}, \citet{yoffe2023statistical},  \citet{buhler2023exploring}, and \citet{faigenbaum2024critical} employed stylometric clustering to detect internal textual divisions corresponding to hypothesized strata—but even these results are subject to interpretative ambiguity, as stylistic boundaries may be blurred by thematic layering or editorial interpolation.

A related example comes from the domain of classical literature. In their stylometric analysis of Latin texts, \citet{Kestemont2016} developed an authorship verification framework to assess the authenticity of disputed works attributed to Julius Caesar. By training classifiers on known Caesar texts and testing against contested pieces, they achieved strong empirical results—but the interpretation of these classifications again hinged on whether the distinguishing signals reflected Caesar’s unique stylistic fingerprint or domain-specific content (e.g., military vocabulary, rhetorical form). While their framework operated under the assumption that thematic bias could be mitigated through feature design and normalization, our approach would enable a more explicit assessment: namely, whether the classification boundaries align with patterns of sequential thematic development rather than true stylistic divergence.

In both historical-critical and classical philological settings, the challenge remains the same: \textit{when classification succeeds, what exactly is being distinguished?} Our hypothesis-testing framework is designed to answer this question directly. By modeling the expected distribution of label sequences that arise from sequentially structured (but non-stylistic) sources, we provide a principled way to separate theme-driven partitions from genuinely stylistic ones. This opens the door for more cautious and informed interpretations of stylometric findings, particularly in cases where authorial boundaries are unknown or the text has been shaped by centuries of redaction, translation, and reuse.

In response to this gap, the present work introduces a hypothesis-testing framework for assessing the influence of sequentially correlated properties—such as theme—on classification outcomes. By generating surrogate label sequences that preserve observed autocorrelation patterns, we estimate whether classification results are likely driven by sequential structure or by non-sequential stylistic signals. This approach complements existing stylometric methods by offering a data-centric, statistically grounded test of interpretability—particularly valuable in contexts where both style and theme may be entangled.

\section{Hypothesis-Testing Scheme} \label{hp_methodology}

We propose a data-centric approach that treats labels as correlated random variables in the sequential domain. This \textit{embedding-invariant} method can be applied to texts (or other datasets) embedded densely, sparsely, or with any subset of features from neural or traditional language models. By estimating the intrinsic average correlations between text units at different lags (i.e., distances along the textual sequence), we generate a multivariate distribution from which we draw label sequences that reflect these sequential correlations. This provides a null hypothesis that simulates the inherent sequential dependencies in the text.

We consider the null hypothesis labeling-generation routine for the simplest scenario of a text composed by one or two authors. 
Once this sampling mechanism for our null distribution is established, we can compute the $p$-value indicating the likelihood that the distinction between the two parts of the text indicated by labels  $L$ is made due to classification driven by sequentially correlated literary properties.

\paragraph{Reference label sequence $L^{(\mathrm{real})}$ of a textual data set.} We consider correlations in the label space. Let $D \in \mathbb{R}^{m \times f}$ denote a textual data set containing $m$ samples (text units -- a sequence of words of some length) of $f$ dimensions (features). We denote a sequence of labels representing some hypothesized or real partition between or within texts as $L^{(\mathrm{real})}$.

\paragraph{Label sequence $L$ of a classification algorithm applied to $D$.}
 We apply some classification algorithm with two labels to $D$ and receive a sequence of labels $L \in \{0,1\}^m$. 

\paragraph{Auto-covariance of $L$.} We proceed to compute the average auto-covariance of $L$ for all lags, to receive a vector $A \in \mathbb{R}^m$, where $A_\ell$ ($\ell=1,\ldots,m$) is the average auto-covariance of $L$ at lag $\ell$, given by
\begin{equation} \label{eq_autocov}
A_\ell = \frac{1}{m}\sum_{j = 1}^{m - \ell}(L_j - \bar{L})(L_{j + \ell} - \bar{L}),
\end{equation}
where $\bar{L} = 1/m \sum_i L_i$ is the average of the elements of $L = (L_1, \ldots, L_m)^T$. To impose linear decay as a function of lag (i.e., that the strength of the correlation between farther separated text units decays linearly), all terms in $A$ are scaled by $1 / m$, rather than the expected term for averages $1 / (m - \ell)$.
In other words, Eq.~\eqref{eq_autocov} is equivalent to 
\[
A_\ell = \frac{m-\ell}{m}\left[\frac{1}{m-\ell}\sum_{j = 1}^{m - \ell}(L_j - \bar{L})(L_{j + \ell} - \bar{L})\right]
,\]
which approaches zero linearly as $\ell\to m$.

\paragraph{Average auto-covariance matrix $M$.} Let $M \in \mathbb{R}^{m \times m}$ denote the average autocovariance matrix, whose entries are defined by
\begin{equation}
(M)_{ij} = A_{|i - j|}.
\label{eq_matrixM}
\end{equation}

Thus, the correlation between every $i$th and $j$th entries in $L$ is given by the (linearly decaying) average autocovariance of $L$ at lag $|i - j|$. 
This approach represents the empirical expectation value of the \textit{global} correlation trend in the text of every two labels at some fixed lag of size $|i - j|$.

\paragraph{Generating a correlated label sequence $L^{(null)}$.} We use the autocovariance matrix $M$ to draw a random sequence of correlated labels $L^{(\mathrm{null})} \in \mathbb{R}^m$ in the following manner: consider the multivariate normal distribution \smash{$V \sim \mathcal{N}(\overrightarrow{\bar{L}}, M)$, where $\overrightarrow{\bar{L}} = (\bar{L},\ldots,\bar{L})^T \in \mathbb{R}^m$}. The choice of $V$ as a multivariate normal distribution stems from the ease of embedding the average autocovariance matrix $M$ as a covariance matrix. Let $X = (X_1, ..., X_m )\in \mathbb{R}^m$ be drawn from $V$, which can be transformed into a vector of correlated labels ($L^{(\mathrm{null})}$) using element-wise thresholding \citep{leisch1998generation}:
\begin{equation*}
L_i^{(\mathrm{null})} =
\begin{cases}
    1, & X_i > 0.5\\
    0, & \textrm{otherwise}.
\end{cases}
\end{equation*}
Essentially, this means that the probability $p_i$ of the $i$th label in $L^{(\mathrm{null})}$ to be 1 is given by
\begin{equation} \label{eq_threshold}
p_i \equiv \mathbb{P}(L^{(\mathrm{null})}_i = 1 ) = \mathbb{P}(X_i > 0.5).
\end{equation}

\paragraph{Intuition.}
The multivariate normal distribution $V$ offers a convenient and analytically tractable method for generating stochastic binary label sequences that preserve the sequential dependencies observed in a reference sequence $L$. This model couples two essential components: (\textbf{1}) the marginal probability of a label being 1, which is assumed to be constant across entries and estimated by the empirical mean of $L$ (i.e., $\bar{L}$), and (\textbf{2}) the average autocovariance matrix $M$, which encodes the expected correlation between each label and every other label in the sequence based solely on their relative distance.

The matrix $M$ is constructed as a symmetric Toeplitz matrix, with entries defined by $(M)_{ij} = A_{|i-j|}$, where $A$ is the vector of average autocovariance across all lags. This Toeplitz structure reflects an important modeling assumption: that the correlation between labels is translation-invariant—that is, the degree of correlation between any two labels depends only on how far apart they are, not on their absolute position in the sequence. This aligns with the hypothesis that thematic or stylistic coherence decays smoothly with distance in natural language, regardless of context or boundary location.

A sample vector $X$ drawn from $V$ is thus a continuous-valued realization that preserves the second-order (i.e., pairwise) sequential correlations of $L$. To convert this to a binary label sequence, we apply element-wise thresholding (e.g., assigning a value of 1 if $X_i > 0.5$ and 0 otherwise). This produces a new binary sequence that shares the same correlation structure as the original, without inheriting any specific content or higher-order dependencies.

Within a hypothesis-testing framework, this allows us to define a null hypothesis: that the observed reference label sequence $L$ reflects nothing more than the natural sequential correlations found in the text. Under this null, $L$ is treated as a representative draw from the multivariate normal model $V$, which is then used to simulate a distribution of surrogate label sequences possessing similar sequential properties.

We then evaluate the degree of agreement between each surrogate sequence and a target label sequence $L^{(\mathrm{real})}$—which may represent a hypothesized or externally assigned partition of the text (e.g., by authorship or stylistic phase). If the agreement between $L$ and $L^{(\mathrm{real})}$ can be explained by sequential correlations alone, then the overlap between the surrogate sequences and $L^{(\mathrm{real})}$ should be statistically comparable. However, if the overlap between $L$ and $L^{(\mathrm{real})}$ significantly exceeds what is expected under the null, we infer that their agreement likely reflects a meaningful, non-sequential structure—such as a real authorial distinction.

\begin{figure*}[t!]  
\centering
\rotatebox[origin=c]{0}{\includegraphics[scale = 0.34]{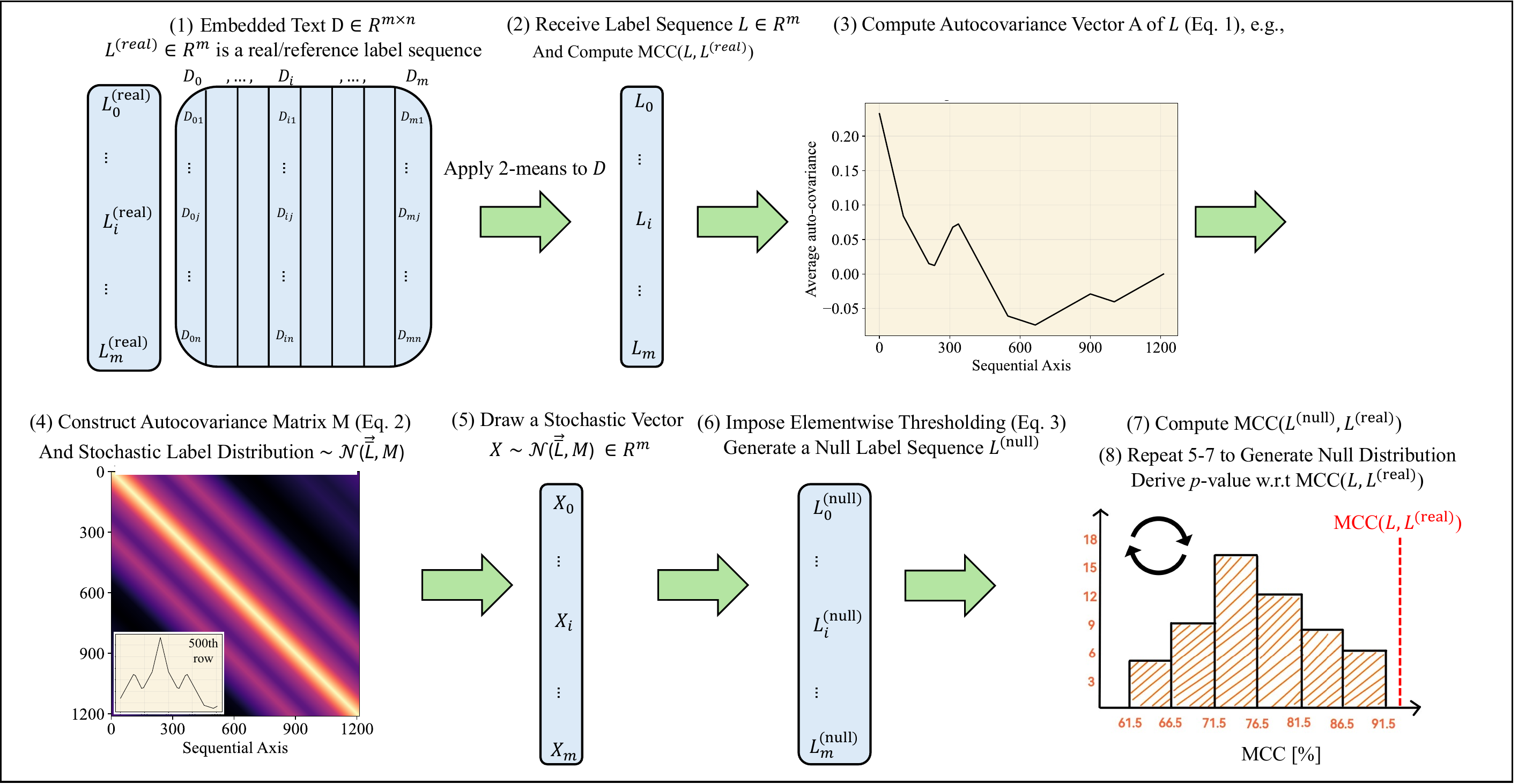}}
\caption{Flowchart of the hypothesis-testing framework. Starting from an embedded text corpus $D \in \mathbb{R}^{m \times n}$, a binary label sequence $L \in \{0,1\}^m$ is obtained via 2-means clustering. The autocovariance vector $A$ and corresponding matrix $M$ are computed from $L$. A stochastic vector $V \sim \mathcal{N}(\vec{\bar{L}}, M)$ is drawn and thresholded to generate null sequences $L^{(\text{null})}$. Repeating this process yields a null distribution of MCC scores against $L$, from which a $p$-value is estimated.}

     \label{Fig_intuition_plot}
\end{figure*}

This approach thus enables a principled test of whether the separability in a classification task is driven by underlying literary structure rather than being an artifact of natural sequential coherence in language.
In Figure~\ref{Fig_intuition_plot}, we visualize the autocovariance vector $A$ and matrix $M$ computed from a label sequence $L$ of length 1213, illustrating the sequential correlation structure modeled in our hypothesis-testing framework.

\paragraph{Matthews correlation coefficient (MCC).} We consider the MCC model evaluation metric, given by
\begin{equation} \label{eq_MCC}
\text{MCC} = \frac{\text{TN} \times \text{TP} - \text{FN} \times \text{FP}}{\sqrt{(\text{TP} + \text{FP})(\text{TP} + \text{FN})(\text{FN} + \text{FP})(\text{TN} + \text{FN})}},
\end{equation}
where T (F) stands for true (false) and P (N) stands for positive (negative), which are evaluations of some sequence against another. 
We use MCC to evaluate the agreement between the reference label sequence $L^{(\mathrm{real})}$ and $L$, or a drawn label sequence $L^{(null)}$ as our statistic of choice, normalized to percent within the range of 50\%--100\%, where 50\% suggests an arbitrary overlap between two label sequences, and 100\% suggests perfect overlap. By drawing multiple random correlated label sequences $L^{(\mathrm{null})}$, we empirically estimate the null distribution of MCC between the random and $L^{(\mathrm{real})}$. We can derive the probability of the MCC score between the latter and $L$ to be drawn from that distribution -- implying that the unsupervised partition is based predominantly on sequentially correlated literary properties. 

We present the procedure of generating a null label sequence $L^{\mathrm{(null)}}$ using this method schematically in Algorithm \ref{alg_ht}.

\begin{algorithm}
\caption{\textbf{Hypothesis-Testing Scheme}: Generating a null label sequence $L^{\mathrm{(null)}}$} 
	\begin{algorithmic}[1]
            \State Apply a classification algorithm on an embedded corpus $D \in \mathbb{R}^{m\times f}$ and receive a label sequence $L \in \mathbb{R}^m$.
            \State Calculate the (decaying) average autocovariance array $A$ of $L$, where $A_\ell$ is computed with Eq. \ref{eq_autocov}, and $\ell \in \{1,\ldots,m \}$.
            \State Generate the average autocovariance matrix $M$, such that $(M)_{ij} = A_{|i - j|}$.
            \State Generate a multivariate normal distribution $V \sim \mathcal{N}(\overrightarrow{\bar{L}}, M)$, where $\overrightarrow{\bar{L}} = (\bar{L},\ldots, \bar{L}) \in \mathbb{R}^m$.
            \State Draw a vector $X = \{X_1,\ldots,X_m \}$ from $V$.
            \State Impose element-wise thresholding on $X$, such that $X_i = 1 \Leftrightarrow X_i > 0.5$, and $X_i = 0$ otherwise.
            \State $X$ is now $L^{\mathrm{\mathrm{(null)}}}$.
	\end{algorithmic} 
\label{alg_ht}
\end{algorithm}

\section{Experiments} \label{experiments}

\subsection{Rationale} \label{experiments_rationale}

Rather than relying solely on state-of-the-art neural language models, we emphasize the value of traditional features that have demonstrated long-standing efficacy in textual analysis. This is particularly relevant for unannotated historical texts, which are often redacted or multilayered, and present stylistic and structural complexities not easily captured by modern embeddings \citep[e.g.,][]{buhler2023exploring}. Frequency-based representations such as tf-idf over word or character $n$-grams remain powerful tools in such contexts.

To complement these, we incorporate a Delta-like embedding scheme inspired by Burrows' Delta \citep{burrows2002delta}, which uses Z-score standardized frequencies of the most frequent $n$-grams or $k$-mers to construct interpretable, content-agnostic stylistic profiles. This representation enhances interpretability and serves as a valuable baseline for clustering and comparison, aligning our pipeline with stylometric tradition.

Our aim is to evaluate how effectively various embeddings capture non-sequentially correlated literary properties. In supervised experiments, we train models to differentiate between text pairs authored either by the same or different individuals. This allows us to examine how supervised systems may encode implicit biases from training labels, potentially reflecting thematic or structural assumptions.

Unsupervised experiments, by contrast, reveal the intrinsic structure of stylistic variation without relying on annotation. By applying clustering to traditional, Delta-style, and neural embeddings, we probe whether modern embeddings offer finer stylistic discrimination or whether classic stylometric indicators remain competitive, especially in genre-diverse or data-scarce contexts.

To further enhance interpretability, we apply a hypothesis-testing framework that quantifies the degree to which classifications rely on sequentially versus non-sequentially correlated features. This enables us to distinguish between classifications driven by authorial style and those influenced by thematic content, offering a deeper understanding of model behavior and interpretive validity.

\subsection{Corpora} \label{experiments_corpora}

We conduct our analysis on several curated English prose corpora selected to represent a broad range of literary styles, genres, and narrative structures. This diversity is designed to stress-test our hypothesis-testing framework under realistic stylometric challenges and to reflect both classical and contemporary authorial voices.

Our datasets are as follows: (
\textbf{1}) the \textit{Harry Potter} series by J. K. Rowling,\footnote{Freely available at \url{https://github.com/formcept/whiteboard/tree/master/nbviewer/notebooks/data/harrypotter}} 
\textbf{2}) four books from the \textit{Percy Jackson} series by Rick Riordan\footnote{Freely available at \url{https://archive.org/details/PercyJacksonTheLightningThief}},
\textbf{3}) works by J.~R.~R. Tolkien, including \textit{The Lord of the Rings} trilogy and \textit{The Silmarillion},\footnote{Freely available at \url{https://archive.org/details/TheSilmarillionIllustratedJ.R.R.TolkienTedNasmith}}
\textbf{4}) selected novels by Charles~Dickens,\footnote{Freely available at \url{https://www.gutenberg.org}}
\textbf{5}) selected works by C.~S.~Lewis,\footnote{Freely available at \url{https://gutenberg.ca/ebooks}}
\textbf{6}) major works by Jane Austen,\footnote{Freely available at \url{https://archive.org/}}
\textbf{7}) detective fiction by Sir Arthur Conan Doyle,\footnote{Freely available at \url{https://sherlock-holm.es/ascii}} and
\textbf{8}) a \textit{spooky} corpus of short fiction by Edgar Allan Poe, H. P. Lovecraft, and Mary Shelley.%
\footnote{Freely available at \url{https://www.kaggle.com/competitions/spooky-author-identification/data}}

We list all texts, their authors, and abbreviations in Table~\ref{app_tab_allTexts} (Appendix \ref{app_text_tabs}). All texts were preprocessed to retain only uncontracted words.\footnote{Preprocessing code available at \url{https://github.com/YoffeG/Thematic-nonThematic_Hypothesis_Testing}} Punctuation was excluded to maintain compatibility with historically noisy or inconsistently punctuated texts, as motivated in Section~\ref{experiments_rationale}.

The corpora span multiple stylistic and thematic regimes: contemporary fantasy (the \textit{Harry Potter} series by J.~K.~Rowling and the \textit{Percy Jackson} series by Rick Riordan), 19th-century literary fiction (works by Charles Dickens and Jane Austen), Christian allegory (\textit{The Screwtape Letters} by C.~S.~Lewis), Gothic and epistolary fiction (\textit{Frankenstein} by Mary Shelley and selected short stories by Edgar Allan Poe), and epic fantasy (\textit{The Lord of the Rings} trilogy and \textit{The Silmarillion} by J.~R.~R.~Tolkien). This enables analyses across: (\textbf{1}) different-author comparisons, (\textbf{2}) intra-author comparisons across genres (e.g., \textit{The Chronicles of Narnia} vs. \textit{The Screwtape Letters}, both by C.~S.~Lewis), and (\textbf{3}) intra-text variation within a single work. Such diversity offers a comprehensive platform for assessing when classification outcomes are thematically or stylistically motivated.

We evaluate 50 randomly selected inter-author pairs and all 55 intra-author permutations. Sampling was stratified by author group, and for each inter-author pair, the number of texts drawn from each author was randomly chosen (up to three), ensuring balanced representation and mitigating genre-specific biases (see Table~\ref{app_tab_textGroups} in Appendix~\ref{app_text_tabs}).

This sampling strategy captures the stylistic range within authors while enabling assessment of stylistic consistency under theme shifts, a key confound in stylometry. By analyzing both inter- and intra-author comparisons, we build a more nuanced picture of how style interacts with genre and structure.

Tables \ref{app_tab_unsupervised_texts_differentAuthors} and \ref{app_tab_unsupervised_texts} (Appendix~\ref{app_text_tabs}) detail the 50 inter-author and 55 intra-author pairs. The latter include genre-similarity annotations, with uncertain cases marked ``?''. These annotations support the interpretation of classification outcomes and help evaluate the method's sensitivity to stylistic versus thematic properties.

\subsection{Preprocessing} \label{preprocessing}

In our preprocessing pipeline, we employ several techniques to standardize and prepare the text data for embedding. First, we convert all texts to lowercase to ensure uniformity and mitigate the impact of case sensitivity on subsequent processing steps. Additionally, we systematically remove punctuation marks to focus solely on the textual content. Following this, we concatenate all the texts into a single continuous string, facilitating the segmentation of the corpus into text units of a desired length. 

In the traditional embedding setup (\S\ref{embedding_traditional}), all texts undergo named entity recognition and removal using spaCy's entity extraction routine,\footnote{All entities apart from the following are removed: CARDINAL, ORDINAL, QUANTITY, PERCENT, TIME, DATE, LANGUAGE, PRODUCT.} \citep{vasiliev2020natural} to decrease the number of features that are of predominantly thematic and semantic merit \citep{rios2017relevance}.

To simulate realistic document mixing, we split the text composed by the second author in the pair into segments, based on a Poisson distribution with $\lambda = 3$, which introduces variability in the length of the segment. These segments are then inserted into arbitrary locations within the first text, rather than concatenated linearly.\footnote{A list of how each pair of texts was commingled can be found at \url{https://github.com/YoffeG/Thematic-nonThematic_Hypothesis_Testing}}
This method enhances the realism of document integration by mimicking potential co-authored document interspersion. 
It avoids oversimplified models and tests the robustness of our approach under various conditions.
A $\lambda$ value of 3 strikes a balance by avoiding excessively short segments that could disrupt coherence, while still allowing larger (and fewer) blocks to be effectively probed.

\subsection{Embedding} \label{embedding}

\subsubsection{Traditional Embedding Setup} \label{embedding_traditional}

We use two complementary feature extraction methods: traditional tf-idf vectorization \citep{aizawa2003information} and a Z-score-based embedding inspired by Burrows’ Delta \citep{burrows2002delta} (see Appendix \ref{app_zscore}). Tf-idf has been shown to perform robustly in unsupervised classification, especially when neural models are unavailable \citep[e.g.,][]{fabien2020bertaa, marcinczuk2021text}. In contrast, the Delta embedding standardizes $n$-gram or $k$-mer frequencies, yielding interpretable stylistic profiles that suppress topical bias. This dual representation enables comparison between raw and normalized frequency-based embeddings. In line with the rest of our experimental setup, we use Z-score standardized vectors in conjunction with $k$-means clustering rather than applying Cosine Delta directly. However, Z-score embeddings yield Euclidean-compatible vectors that can be meaningfully averaged, making them a methodologically coherent choice for unsupervised clustering within our pipeline \citep{evert2015towards}.

Each text pair is embedded using parameter grids over $n$ and $\ell$: word $n$-grams with $n \in {1,2,3,4}$, character $k$-mers with $n \in {1,2,3,4,5,6}$, and text unit lengths $\ell \in {10, 50, 100, 250, 500, 750, 1000, 2000}$. These ranges span fine- to coarse-grained representations and allow us to probe stylistic patterns across linguistic levels \citep[e.g.,][]{antonia2014language}. Increasing $n$ in word $n$-grams captures syntactic structure, while character $k$-mers track morphology and orthography. Varying $\ell$ helps us assess stability from sentence level to discourse level.

For both tf-idf and Z-score embeddings, we select the top $f \in {100, 200, 300}$ most frequent features across the joint corpus. This reduces noise from rare or content-specific tokens \citep{akiva2012identifying}. Importantly, the same feature set is used for both embeddings to ensure comparability. Focusing on shared high-frequency features improves signal-to-noise ratio and reduces sparsity—a known challenge in stylometric modeling \citep{lagutina2019survey}.

To ensure robustness, we apply consistent subsampling: a random subset of $c \cdot \min{m_1^{(\ell)}, m_2^{(\ell)}}$ text units is selected per document, with $c = 0.2$ fixed across all settings. This choice balances computational cost and classification stability based on preliminary tests of MCC variance. Evaluating each $(n, \ell, f)$ configuration across both embeddings provides a rigorous framework for analyzing the interplay between feature space design and stylistic classification.

\subsubsection{Neural Embedding Setup} \label{embedding_neural}

We use an off-the-shelf implementation of the STAR\footnote{Available at: \url{https://huggingface.co/AIDA-UPM/star}} (Style Transformer for Authorship Representations) embedding \citep{Huertas-Tato2023Oct} to test the validity of our approach when using a state-of-the-art authorship-attribution-oriented neural embedding. STAR is a contrastively trained model fit to learn authorship embeddings instead of semantics. 
Similarly to the traditional embeddings, we implement STAR in two scenarios: 
(\textbf{1}) The unsupervised approach (\S\ref{classification_us}), whereby we cluster texts using a $2$-means clustering algorithm on STAR-embedded pairs of texts and apply our hypothesis-testing framework to that classification. 
(\textbf{2}) A supervised cosine similarity approach (\S\ref{classification_s}), as implemented in \citet{Huertas-Tato2023Oct}. In this case, a comparison between the traditional style-bearing features and STAR emphasizes the potential advantage novel neural models that are optimized to preserve and isolate the stylistic signal may have therein.

\subsection{Classification} \label{classification}

\subsubsection{Unsupervised Approach} \label{classification_us}

For our unsupervised experiments, we adopt $k$-means clustering with $k = 2$—a standard and interpretable algorithm compatible with our embedding choices and suited to our binary classification framework. Each pair of texts is embedded using varying combinations of word $n$-gram or character $k$-mer sizes and text unit lengths, producing a feature matrix $T^{(\ell, n, f)}$ where $n$, $\ell$, and $f$ represent the token granularity, text unit length, and number of selected features, respectively. $k$-means clustering is then applied to this matrix to generate cluster labels, which are evaluated using the Matthews correlation coefficient (MCC) against the known binary labels of commingled text segments.

To ensure statistical robustness and reduce dependence on specific text partitions, we apply a batch-sampling strategy. For each parameter configuration, we draw 20 random subsamples per experiment. Each subsample includes $c \cdot \min{m_1^{(\ell)}, m_2^{(\ell)}}$ text units from each document, with $c = 0.2$ selected empirically to balance variance and computational cost. We randomly sample unordered text units and then sort them back into their original sequence order (using \texttt{argsort}) to preserve the autocorrelative structure. This avoids biases introduced by local anomalies such as shifts in topic or narrative style.

Each batch subsample is clustered using $k$-means, producing a predicted label sequence $L$. We then compare $L$ to the true partition $L^{\mathrm{(real)}}$ using MCC to assess alignment. To evaluate whether such alignment arises from intrinsic stylistic divergence rather than from sequential correlation alone, we apply our hypothesis-testing framework (\S\ref{hp_methodology}). Specifically, we generate a null distribution of MCC scores using 1000 synthetic labelings $L^{\mathrm{(null)}}$ that match the autocorrelation structure of $L$ but are otherwise random. The $p$-value for each observed MCC is computed relative to this distribution, quantifying the probability of obtaining such a result under the null hypothesis of sequential correlation alone.

As part of our embedding design, we include both tf-idf vectorization and a Z-score–based embedding inspired by Burrows’ Delta \citep{burrows2002delta} (Appendix~\ref{app_zscore}). The Z-score embedding uses standardized frequency profiles of the most common $n$-grams or $k$-mers across the corpus. This approach emphasizes deviations from corpus-level norms, yielding stylistically interpretable feature vectors. While not explicitly designed for supervised classification, such embeddings have demonstrated strong performance in unsupervised authorship attribution tasks and are particularly effective when paired with distance-based methods such as clustering \citep{evert2015towards}. \citet{evert2015towards} show that Delta-style embeddings, especially when paired with normalization strategies, can achieve competitive clustering results without supervision. These properties make Z-score embeddings well-suited as an unsupervised baseline for evaluating stylistic separability, particularly in contexts where interpretability and content-agnostic analysis are essential. Moreover, since $k$-means operates in Euclidean space, Z-score embeddings—unlike Cosine Delta—are naturally compatible with centroid-based clustering, as discussed in \S\ref{embedding_traditional}.

After completing batch sampling and clustering, we compute the empirical distributions of MCC scores and corresponding $p$-values for each parameter combination. To address the issue of multiple hypothesis testing, we apply the Benjamini-Hochberg false-discovery rate (FDR) procedure \citep{benjamini1995controlling} across all $(n, \ell)$ settings. The resulting average MCC scores and FDR-corrected $p$-values are used as summary statistics to interpret classifier performance.

To quantify the stability of each test result, we compute a 95\% Wilson score confidence interval for the unadjusted $p$-value, before FDR correction. For all reported significant results ($p < 0.05$), we confirm that the upper bound of this interval also falls below 0.05. While this interval applies only to unadjusted $p$-values, its use supports a more conservative and stable interpretation of statistical significance prior to correction.

\subsubsection{Supervised Approach (General Imposters/Cosine Similarity Methods)}  \label{classification_s}

Here, our goal is to test the susceptibility of supervised classification methods to false-positive classifications of texts that are attributed to, for example, differences in authorial style, while being indeed affected by sequentially correlated literary properties, such as thematic content.
This susceptibility arises because the chosen subsets of features often encapsulate multiple facets of the text, such as vocabulary, syntax, and punctuation, leading to a blending of stylistic and genre-wise characteristics within the classification process. Moreover, traditional features are not optimized to represent some desired literary property; rather, they are selected based on theoretical frameworks or limited empirical findings, potentially overlooking nuanced stylistic distinctions.
We consider two supervised classification approaches: (\textbf{1}): The General Imposters\footnote{We used the version available at \url{https://github.com/bnagy/ruzicka}, originally developed within the scope of the work of \citet{Kestemont2016}.} (GI) framework \citep{koppel2014determining} -- a gold-standard supervised method used in multiple authorship-attribution scenarios \citep[e.g.,][]{ juola2021verifying}. 
The GI model is designed to account for the possibility that authors might mimic or imitate the writing styles of others, either intentionally or unintentionally. This means the model doesn't just rely on a single set of features to identify an author's style but rather considers multiple feature subsets that capture various aspects of writing style. 
(\textbf{2}): Cosine similarity. In this approach, as implemented in \citet{Huertas-Tato2023Oct}, classification is performed by attributing an embedded text unit to either one or another centroid of a training-set sample of each class (using the original partition $L^{\mathrm{real}}$) by measuring the cosine similarity between them. In this case, all features are used to compute the cosine similarity.

Neural embeddings, unlike traditional features such as word $n$-grams, inherently capture diverse linguistic properties without manual selection. Attempting to under-sample or select subsets of features from neural embeddings, as in the GI framework, may not be suitable due to their holistic representation of text. Utilizing all features in neural embeddings is often more effective for classification, leveraging their rich linguistic representations encoded in high-dimensional spaces. 
Therefore, in our analysis, we apply the GI framework to classify the traditionally-embedded texts, whereas in the neurally-embedded case, we apply the cosine similarity classification. We set a consistent train/test split ratio of 20\%--80\% in all supervised scenarios to evaluate the model's performance and assess the model's generalization ability without favoring either the training or test.

The supervised classification procedure using the GI and cosine similarity frameworks for some parameter combination of $n$ (only in the GI case) and $\ell$ and a feature space $f$ is summarized in Algorithms 2 and 3 in Appendix B, respectively.

\section{Results} \label{results}

Here, we present the results of experiments where we (\textbf{1}) distinguish 50 pairs of texts composed by different authors and all (55) permutations of pairs of texts written by the same author in our corpus, and (\textbf{2}) apply our hypothesis-testing method on all resulting classifications to determine whether it is based on predominantly sequentially correlated literary properties or not. 
We perform these experiments in four configurations, where in all cases, we split the texts into units of varying length (\S\ref{embedding}):

\begin{trivlist}

    \item[~$\bullet$] \textbf{Unsupervised Classification of Traditionally-Embedded Texts}: We apply the $2$-means algorithm (\S\ref{classification_us}) to classify texts embedded according to either word $n$-grams or character $k$-mers of varying sizes and numbers of corpus-wise-weighed most-frequent features (\S\ref{embedding_traditional}).

    \item[~$\bullet$] \textbf{Supervised Classification of Traditionally-Embedded Texts}: We apply the supervised GI framework (\S\ref{classification_s}) to classify texts embedded according to either word $n$-grams or character $k$-mers of varying sizes and numbers of corpus-wise-weighed most-frequent features (\S\ref{embedding_traditional}).

    \item[~$\bullet$] \textbf{Unsupervised Classification of Neurally-Embedded Texts}: We apply $2$-means  (\S\ref{classification_us}) to classify texts embedded using the pretrained STAR model (\S\ref{embedding_neural}).

    \item[~$\bullet$] \textbf{Supervised Classification of Neurally-Embedded Texts}: We apply the supervised cosine similarity framework (\S\ref{classification_s}) to classify texts embedded using the pretrained STAR model (\S\ref{embedding_neural}).

\end{trivlist}

In Figure~\ref{Fig_res_words}, we demonstrate the output of one such experiment for the case of 50 pairs of texts composed by distinct authors, embedded using word $n$-grams with $f = 300$ features, and classified using the unsupervised 2-means clustering approach. The x-axis represents all tested parameter combinations of $n$-gram size ($n$) and text unit length ($\ell$), while the y-axis corresponds to the 50 evaluated text pairs. Each cell in the matrix is color-coded by the normalized Matthews correlation coefficient (MCC) score for that pair under the given $(n, \ell)$ configuration, but only if the classification was found to be statistically significant under our hypothesis-testing framework (i.e., $p < 0.05$  after FDR correction). Cells where the classification was not deemed significant are left blank.

This visualization highlights parameter regimes where stylistic signals dominate over sequential correlations, as evidenced by high MCC scores passing the hypothesis test. Importantly, we observe that certain configurations—particularly smaller $n$-grams and moderate unit lengths (e.g., $n = 1$--$2$, $\ell = 100$--$500$)—tend to yield more consistent success in distinguishing authorial style across text pairs. Conversely, longer text unit lengths or higher $n$-gram values often lead to sparsity and reduced classification power, reflected in fewer significant cells.

The blank entries serve as a visual cue for parameter settings where classification success was likely driven by sequentially correlated properties (e.g., theme or narrative flow), as such cases fail to reject the null hypothesis. This reinforces the utility of our method in diagnosing when classification is style-driven versus thematically biased.
We note that results for smaller feature set sizes ($f = 100, 200$) were qualitatively similar to those of $f = 300$, and are therefore omitted for clarity.

\begin{figure}
\centering
\rotatebox{90}{\includegraphics[scale = 0.09]{words_diff_kmeans-batch-0.2.pdf}}
  \caption{Significance map for applying 2-means classification to 50 pairs of commingled texts by different authors (see Table A.2 in Appendix 1), embedded using word $n$-grams with $f$ = 300. The x-axis shows all feature combinations ($n$ and $l$), and the y-axis lists 50 text pairs. Colored cells indicate parameter combinations yielding classifications predominantly affected by non-sequentially correlated properties with high statistical significance, color-coded by normalized MCC score. Blank cells denote classifications where the hypothesis test yielded a $p$-value $>$ 0.05.}
     \label{Fig_res_words}
\end{figure}

We report performance using the following metrics. A \textbf{positive} indicates a classification driven by non-sequentially correlated properties (i.e., beyond what is expected under the null), while a \textbf{negative} indicates no such classification. \textbf{True} and \textbf{false} refer to whether the result aligns with ground truth (i.e., different or same author). \textbf{Binary} metrics refer to the presence of at least one significant parameter setting per text pair, while \textbf{total} metrics account for all parameter combinations.


In Tables \ref{tab_word_ngrams}-\ref{Tab_results_rates_star}, we list all true/false positive rates for each experimental setup of traditionally- and neurally-embedded texts. Below, we discuss our results for the traditionally- and neurally-embedded texts.
In Appendices C and D, we provide graphic representations of these experiments.

\begin{table}[h]
\footnotesize
    \centering
\begin{tabular}{|l|l|l|l|l|}
    \hline\hline
   & ($k$-means) Positive & ($k$-means) Negative & (GI) Positive & (GI) Negative  \\
   \hline (Binary) True & \cellcolor{yellow} 100\% & \cellcolor{yellow} 70.9\% (74.5\%) & \cellcolor{pink} 100\%  & \cellcolor{pink} 0.0\% (0.0\%)  \\
         \hline
      (Binary) False &\cellcolor{yellow} 29.1\% (25.5\%) &\cellcolor{yellow} 0.0\% & \cellcolor{pink} 100\% (100\%) & \cellcolor{pink} 0.0\% \\
         \hline
   (Total) True & \cellcolor{green} 44.0\% & \cellcolor{green} 92.1\% (94.9\%) & \cellcolor{lime} 77.6\% & \cellcolor{lime} 66.0\% (68.4\%)  \\
         \hline
   (Total) False & \cellcolor{green} 7.9\% (5.1\%) & \cellcolor{green} 56.0\% & \cellcolor{lime} 44.0\% (42.6\%) & \cellcolor{lime} 22.4\% \\
         \hline
    \end{tabular}
    \caption{Confusion matrices for \textbf{word $n$-gram} embedded texts for the unsupervised ($k$-means) and supervised (GI) classification approaches, with $f$ = 300 (see \S\ref{embedding_traditional}). True positives and false negatives refer to attempting to distinguish pairs of texts composed by different authors. In contrast, false positives and true negatives refer to attempting to distinguish pairs of texts composed by the same author. The bracketed rates indicate the results when overlooking false-positive classifications of texts of explicitly different genres, which we consider to be: \texttt{[Lotr\_1, Silmarillion]}, \texttt{[Lotr\_2, Silmarillion]}, \texttt{[Lotr\_3, Silmarillion]}, \texttt{[Kipling\_JungleBook, Kipling\_Ballads]}, \texttt{[Kipling\_JungleBook2, Kipling\_Ballads]}, \texttt{[Narnia, Screwtape]}, \texttt{[Caspian, Screwtape]} (amounting to 16\% of all pairs of texts composed by the same author in our corpus).}
    \label{tab_word_ngrams}
\end{table}

\begin{table}[h]
\footnotesize
\centering
\begin{tabular}{|l|l|l|l|l|}
\hline\hline
 & ($k$-means) Positive & ($k$-means) Negative & (GI) Positive & (GI) Negative \\
\hline
(Binary) True & \cellcolor{yellow} 98.0\% & \cellcolor{yellow} 69.1\% (79.2\%) & \cellcolor{pink} 100\% & \cellcolor{pink} 10.9\% (12.5\%) \\
\hline
(Binary) False & \cellcolor{yellow} 21.9\% (18.8\%) & \cellcolor{yellow} 2.0\% & \cellcolor{pink} 89.1\% (87.5\%) & \cellcolor{pink} 0\% \\
\hline
(Total) True & \cellcolor{green} 58.9\% & \cellcolor{green} 89.0\% (95.1\%) & \cellcolor{lime} 93.4\% & \cellcolor{lime} 59.1\% (66.1\%) \\
\hline
(Total) False & \cellcolor{green} 11.0\% (4.9\%) & \cellcolor{green} 34.9\% & \cellcolor{lime} 40.9\% (33.9\%) & \cellcolor{lime} 31.1\% \\
\hline
\end{tabular}
\caption{Confusion matrices for \textbf{character $k$-mer} embedded texts for the unsupervised ($k$-means) and supervised (GI) classification approaches, with $f$ = 300, similarly to Table \ref{tab_word_ngrams}.}
\label{tab_char_kmers}
\end{table}

\begin{table}[h]
\footnotesize
\centering
\begin{tabular}{|l|l|l|l|l|}
\hline\hline
 & ($n$-grams) Positive & ($n$-grams) Negative & ($k$-mers) Positive & ($k$-mers) Negative \\
\hline
(Binary) True & \cellcolor{yellow} 100\% & \cellcolor{yellow} 0\% (0\%) & \cellcolor{pink} 100\% & \cellcolor{pink} 70.9\% (81.2\%) \\
\hline
(Binary) False & \cellcolor{yellow} 100\% (100\%) & \cellcolor{yellow} 0\% & \cellcolor{pink} 29.1\% (18.8\%) & \cellcolor{pink} 0\% \\
\hline
(Total) True & \cellcolor{green} 50.5\% & \cellcolor{green} 81.5\% (83.9\%) & \cellcolor{lime} 61.7\% & \cellcolor{lime} 88.7\% (94.8\%) \\
\hline
(Total) False & \cellcolor{green} 19.5\% (16.1\%) & \cellcolor{green} 49.5\% & \cellcolor{lime} 11.3\% (5.2\%) & \cellcolor{lime} 38.3\% \\
\hline
\end{tabular}
\caption{Confusion matrices for Z-score-based unsupervised classification of texts using word $n$-gram and character $k$-mer embeddings, with $f$ = 300, similarly to Table~\ref{tab_word_ngrams}.}

\label{tab_zscore}
\end{table}

\begin{table}[h]
\footnotesize
\centering
\begin{tabular}{|l|l|l|l|l|}
\hline\hline
 & ($k$-means) Positive & ($k$-means) Negative & (cosine) Positive & (cosine) Negative \\
\hline
(Binary) True & \cellcolor{yellow} 100\% & \cellcolor{yellow} 61.8\% (70.8\%) & \cellcolor{pink} 92.0\% & \cellcolor{pink} 60.0\% (68.8\%) \\
\hline
(Binary) False & \cellcolor{yellow} 38.2\% (29.2\%) & \cellcolor{yellow} 0.0\% & \cellcolor{pink} 30.9\% (25\%) & \cellcolor{pink} 8.0\% \\
\hline
(Total) True & \cellcolor{green} 100\% & \cellcolor{green} 65.8\% (75.3\%) & \cellcolor{lime} 85.3\% & \cellcolor{lime} 64.2\% (72.9\%) \\
\hline
(Total) False & \cellcolor{green} 34.2\% (24.7\%) & \cellcolor{green} 0.0\% & \cellcolor{lime} 35.8\% (27.1\%) & \cellcolor{lime} 14.7\% \\
\hline
\end{tabular}
\caption{Confusion matrices for STAR embedded texts for the unsupervised ($k$-means) and supervised (cosine) classification approaches, similarly to Table \ref{tab_word_ngrams}.}
\label{Tab_results_rates_star}
\end{table}

\subsection{Results: Traditionally-Embedded Texts} \label{results_trad}

An intriguing observation arises from the results, highlighting the remarkable true positive rates achieved by unsupervised methods, particularly notable in traditionally-embedded texts. This finding holds significance as unsupervised approaches, such as the $k$-means algorithm, operate without explicit optimization for authorship attribution tasks. Despite this, they exhibit robust performance in identifying texts by the same author, suggesting that these methods capture intrinsic textual similarities effectively. This success underscores the potential of unsupervised techniques in authorship attribution tasks, especially in scenarios where text representations are not specifically tailored for such purposes. As we demonstrate below, further exploration of these unsupervised methods could offer valuable insights into textual data's underlying structures and patterns, enhancing our understanding of authorial styles and linguistic nuances.

Another notable aspect of the results is the relatively low false positive rate observed in the unsupervised method, particularly evident in traditionally-embedded texts. Despite their inherent lack of optimization for authorship attribution, unsupervised methods exhibit the ability to minimize misclassifications of texts by different authors as authored by the same individual. This finding suggests that unsupervised approaches are adept at discerning subtle textual differences that distinguish authors, contributing to their effectiveness in authorship attribution tasks.
That said, it is evidently a nontrivial task to discern between cases where false positive classifications are made based on genre-wise differences or not, as is demonstrated in our results.

In contrast, the supervised GI method demonstrates a strong true positive identification capability, but its performance is marred by high false positive rates. This issue highlights a significant challenge faced by supervised classification approaches, where the classifier erroneously identifies texts by the same author as authored by different ones. Such false positives can be attributed to various factors, including the complexity of linguistic features captured by the embedding techniques and the classifier's susceptibility to common stylistic elements across an author's works.

\begin{figure}
\centering
\rotatebox[origin=c]{0}{\includegraphics[scale = 0.6]{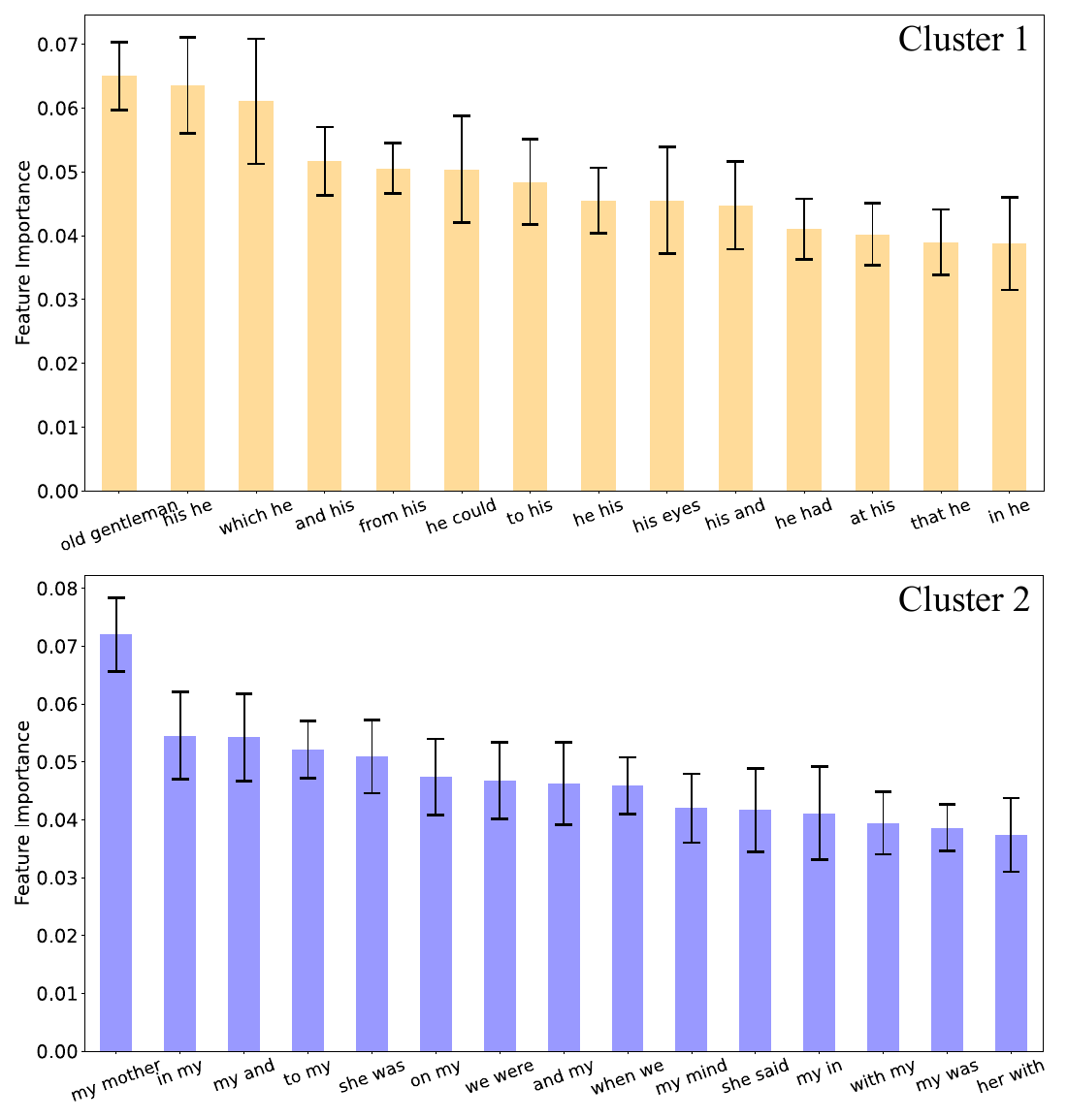}}
  \caption{Extracted important features for the classification of the texts pair \texttt{[Dickens\_Copperfield, Dickens\_OliverTwist]}, embedded using word $n$-gram size 2 with text unit length of 1000.}
     \label{Fig_clusters_Dickens}
\end{figure}

In Figure \ref{Fig_clusters_Dickens}, we apply the important features extraction method introduced in \citet{yoffe2023statistical} on the false positive classification between \texttt{[Dickens\_Copperfield, Dickens\_OliverTwist]}, embedded using word $n$-gram size 2 with text unit length of 1000. This embedding yields an average MCC score of $\approx$90\% and an average $p$-value of $\approx0.01$ -- rendering it a classification that is not driven by sequential correlations with high statistical significance, despite the two texts having both been composed by Charles Dickens. In this case, as demonstrated in Figure \ref{Fig_clusters_Dickens}, the distributions between first- and third-person speech features vary significantly, essentially a difference in genre stemming from the fact that David Copperfield is composed as a semi-auto-biography, leading to the false positive classification.
Thus, by assessing the features yielding the classification -- a considerably less challenging task in the case of traditional embeddings, such as word $n$-grams -- interpretability analysis can provide valuable insight into the reason for false positive classifications and aid in tuning relevant feature subsets to avoid or detect them.

Further insight emerges when comparing tf-idf-based embeddings with those generated using the Z-score framework. As shown in Table~\ref{tab_zscore}, z–score–based unsupervised classification achieves moderately strong true positive rates but slightly underperforms relative to tf-idf, particularly in managing false positives. While both approaches rely on the same set of top 300 most frequent features, the key distinction lies in how these features are encoded. In tf-idf, feature values are weighted by their relative frequency across documents, suppressing tokens that appear uniformly throughout the corpus and emphasizing those that are discriminative but not ubiquitous. This weighting scheme attenuates noise from common tokens and boosts stylistic signals, ultimately improving classification precision. In contrast, Z-score embeddings reflect raw frequency deviations from expected counts, which may be more vulnerable to fluctuations due to topic or unit length, especially in sparse settings. This difference in encoding—not in feature selection—likely explains tf-idf’s superior performance in capturing authorial style with fewer spurious signals.

A notable artifact of the Z-score–based approach is the complete absence of binary true negatives in the $n$-gram case—that is, none of the same-author pairs are deemed stylistically indistinguishable across any tested configuration. This effect is especially concentrated in high-order $n$-grams (e.g., $n = 3,4$) paired with short text unit lengths ($\ell = 100, 250$), as shown in Figure~\ref{Fig_delta_same_ngrams} in Appendix \ref{app_trad}. In such cases, the resulting feature space becomes highly sparse: long $n$-grams are unlikely to repeat across short units, even within texts by the same author. Consequently, minor topical or structural differences can artificially inflate statistical dissimilarity, leading to false positive classifications. In contrast, character $k$-mers exhibit much higher robustness in the same Z-score framework—reflected in the presence of a substantial number of binary true negatives (70.9\%, rising to 81.2\% after genre adjustment). This discrepancy is explained by the inherently denser nature of character $k$-mers, which are more likely to recur across units even in short spans, enabling them to better preserve shared stylistic signals. These findings collectively underscore the importance of choosing features that balance discriminability with statistical stability—an objective well-served by tf-idf weighting and character-level representations.

Our findings provide insight into the widespread effectiveness of various traditional embeddings in distinguishing between text units of varying lengths.

\subsection{Results: Neurally-Embedded Texts} \label{results_star}

Here, an interesting result begs attention: in the unsupervised classification of pairs of text by different authors, neurally embedded texts achieve outstanding performance in binary true and total positive identification rates. In the supervised cosine case, however, both rates drop dramatically. Indeed, the expectation for neural embeddings coupled with supervised classification to outperform all other configurations proved false within our analysis's scope and merits further exploration. 

In contrast to the traditionally-embedded texts scenario, the low false positive rates demonstrate that neural embeddings, such as STAR, can isolate the desired literary component (in this case -- stylistic) and minimize the effect of undesirable influences such as those of thematic content. These results indicate the effectiveness of neural models tailored to the authorship-attribution task in minimizing misclassifications of texts by the same author as authored by different authors, even in supervised classification approaches.

Comparing these results with the performance of supervised methods in traditionally-embedded texts, we observe a considerable disparity in false positive rates. While traditionally-embedded texts presented challenges for supervised methods, leading to relatively high false positive rates, neurally-embedded texts showcase the ability of supervised methods to achieve lower false positive rates. This discrepancy suggests that the inherent characteristics of neurally-embedded representations may offer advantages in mitigating false positives, thereby enhancing the reliability and accuracy of supervised authorship attribution systems. Further investigation into the underlying mechanisms driving these differences could yield valuable insights into the optimization and refinement of authorship attribution techniques across text embedding methodologies.

One limitation inherent in the exploration of neurally-embedded texts is the challenge of interpretability, which restricts the depth of analysis, particularly regarding phenomena such as false positives. Unlike traditional embedding methods, which often provide explicit linguistic features that can be examined and manipulated to understand model behavior, neural embeddings operate more abstractly, making it challenging to dissect the underlying representations comprehensively. Consequently, while we could delve into false positives and their implications in the context of traditionally-embedded texts, a similar level of interpretability may not be achievable with neural embeddings. This limitation underscores the trade-off between the richness of representation offered by neural embeddings and the interpretability required for in-depth analysis, highlighting the need for innovative approaches to bridge this gap and unlock the full potential of neurally-embedded representations in authorship attribution tasks.

\subsection{Comparative Discussion}

The central goal of this work is to evaluate the extent to which textual classification is influenced by sequentially correlated properties (e.g., thematic continuity), as opposed to non-sequential features such as stylistic markers, across a range of corpora and methods. Our experimental design—spanning both supervised and unsupervised classification, as well as traditional and neural embeddings—was applied to corpora with clearly attributed authorship. Here we discuss cross-method and cross-corpus trends observed in the results.

Our experiments show that unsupervised $k$-means clustering using traditional tf-idf-based embeddings (word $n$-grams and character $k$-mers) consistently yields high true positive rates when distinguishing between texts by different authors. This is especially evident in author pairs with distinct genres and vocabularies (e.g., Charles Dickens vs. J.~R.~R.~Tolkien, Jane Austen vs. H.~P.~Lovecraft). Importantly, our hypothesis-testing framework confirms that many of these classifications are driven by non-sequential properties, indicating a robust capture of authorial style.

False positives in unsupervised traditional embeddings are largely restricted to intra-author comparisons involving genre variation. A notable case is \textit{David Copperfield} vs. \textit{Oliver Twist} by Charles Dickens, where stylistic misclassification likely results from narrative mode differences (first-person vs. third-person), as visualized in Figure~\ref{Fig_clusters_Dickens}. This example demonstrates the sensitivity of unsupervised methods to intra-author variation, particularly when accompanied by structural or genre shifts.

Supervised classification using the General Imposters framework achieves strong true positive performance but exhibits high false positive rates in same-author comparisons. This behavior is likely due to its reliance on frequent feature subsets, which may be overly responsive to topical or genre-specific elements. As such, it can overfit superficial differences between same-author texts. Our hypothesis-testing procedure helps detect these misclassifications by quantifying whether they reflect genuine stylistic divergence or merely thematic shifts.

The Z-score embedding results offer additional insight. While the Z-score–based approach performs slightly below tf-idf in terms of true positive rates, particularly for word $n$-grams, it yields interpretable and stylistically grounded embeddings. Notably, Z-score classification exhibits high false positive rates for same-author word $n$-gram comparisons—largely driven by feature sparsity at higher $n$ and shorter text units. This sparsity leads to exaggerated stylistic differences between texts that share a genre or topic but diverge structurally. In contrast, Z-score embeddings using character $k$-mers perform substantially better in terms of binary true negatives, confirming their robustness in same-author settings and their lower sensitivity to sparsity.

Neural embeddings (STAR) present a nuanced pattern. In unsupervised classification, they deliver strong true positive rates with relatively low false positives. However, supervised classification using cosine similarity shows a marked increase in false positives—particularly for genre-divergent works by the same author (e.g., C.~S.~Lewis's \textit{The Chronicles of Narnia} vs. \textit{The Screwtape Letters}). This suggests that even embeddings optimized for authorial style can conflate stylistic and thematic dimensions when coupled with supervised similarity measures.

Across all methods, authors like J.~K.~Rowling and Jane Austen show consistent intra-author classification, reflecting stable stylistic patterns. By contrast, authors like Rudyard Kipling, Charles Dickens, and J.~R.~R.~Tolkien demonstrate greater intra-author variability, consistent with their genre-spanning outputs. These patterns highlight the importance of contextualizing classification outcomes with genre and narrative structure in mind.

Overall, unsupervised approaches using tf-idf and Z-score embeddings—particularly with character $k$-mers—yield the most interpretable and stable results, combining high true positive rates with comparatively low false positives. While supervised methods offer stronger discriminative power, they are more vulnerable to misclassification in the presence of genre or topic shifts. Neural embeddings are competitive, but less interpretable and more sensitive to genre effects. In all cases, our hypothesis-testing framework plays a central role in disentangling whether classification decisions reflect genuine stylistic distinction or are confounded by sequential correlations.

\begin{figure}[t!]
\hspace*{-1.2cm} 
\centering
\rotatebox[origin=c]{0}{\includegraphics[scale = 0.5]{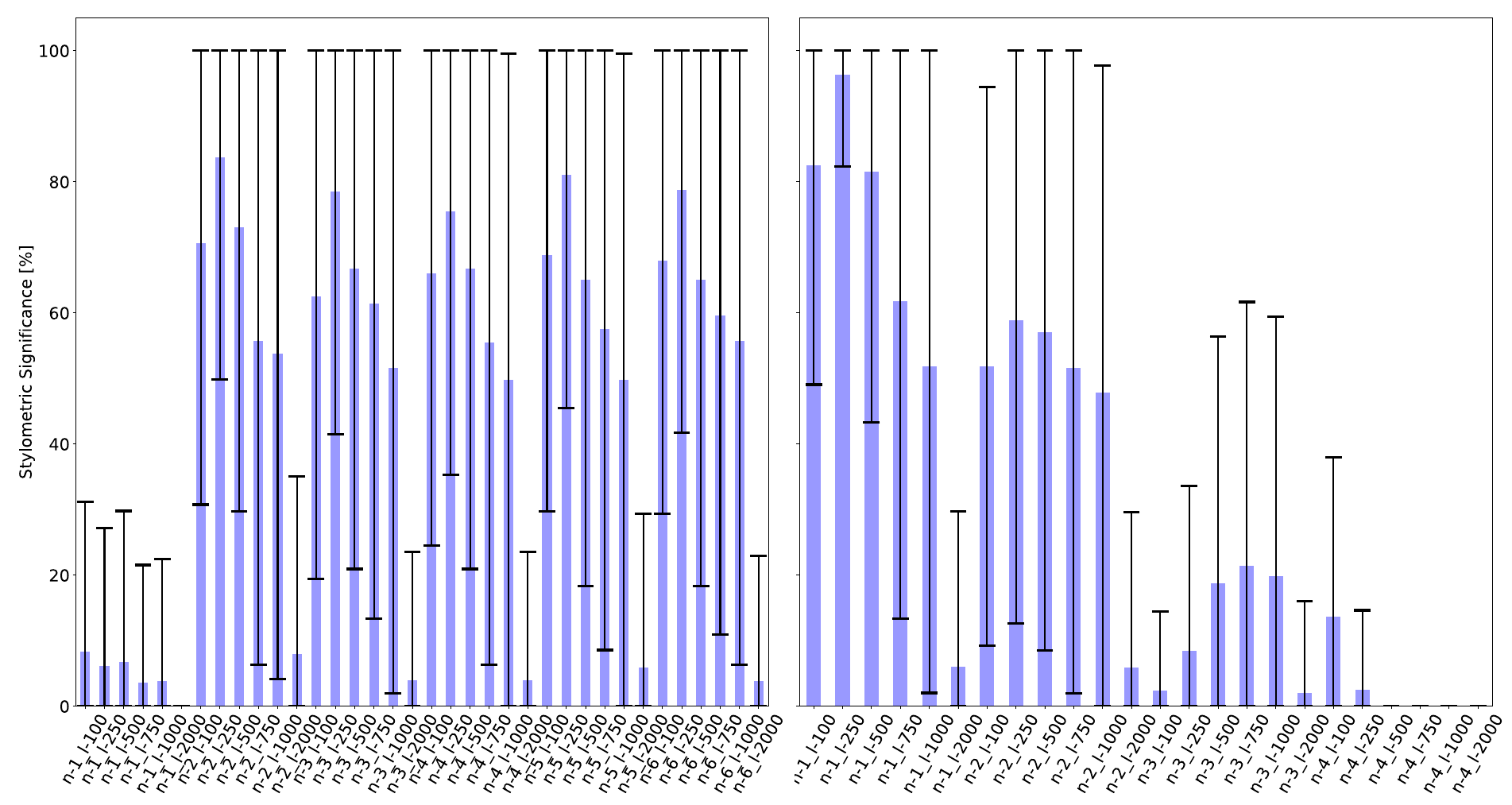}}
\caption{Average distinguishing power of traditional feature configurations across all true positive cases. Each value reflects the mean and standard deviation of the MCC scores over text pairs composed by different authors. \textbf{Left Panel:} Character $k$-mers ($k = 2$ to $6$). \textbf{Right Panel:} Word $n$-grams ($n = 1$ to $4$).}
\label{Fig_stylometric significance}
\end{figure}

Lastly, we assess the stylometric strength of traditional feature configurations. Figure~\ref{Fig_stylometric significance} reports MCC scores across all true positive classifications. Character $k$-mers perform robustly across all $k$ values, especially with shorter text units. Unlike word $n$-grams, they avoid the sparsity problem at higher orders and remain informative even at $k = 6$. Word $n$-grams, by contrast, exhibit declining performance with increasing $n$, as longer $n$-grams introduce sparsity and reduce the consistency of stylistic capture. These findings emphasize the value of short $n$-grams and character-level features in producing dense, reliable stylistic signals.

\section{Conclusions and Limitations}

We propose a novel method for modeling sequential correlations in texts by encoding the dependencies of a label sequence into a correlated multivariate normal distribution. This enables the stochastic generation of label sequences that preserve the sequential structure of the original data, forming the foundation of a hypothesis-testing framework designed to assess whether observed classification results are primarily driven by these correlations.

We evaluate this method on English prose texts—both by the same and different authors—to distinguish classifications influenced by sequentially correlated literary properties, such as thematic content, from those driven by non-sequential features like stylistic variation. Using traditional word $n$-grams, character $k$-mers, and neural embeddings optimized for authorial style, we apply both supervised and unsupervised classification methods. Our results show that the proposed approach effectively identifies theme-driven classification behavior and enhances interpretability in stylometric tasks.

We find that unsupervised methods—particularly those using traditional embeddings—perform remarkably well in distinguishing texts by different authors, yielding high true positive rates and low false positives. Neural embeddings, while powerful in distinguishing between authors, sometimes misclassify texts by the same author when those texts differ in genre, suggesting a degree of sensitivity to thematic or structural variation. Supervised methods show higher susceptibility to false positives, especially when relying on feature sets that blend thematic and stylistic cues—underscoring the value of interpretability and the importance of disentangling confounding influences.

Performance is also modulated by text unit length: character $k$-mers tend to yield consistent results across unit sizes, whereas the effectiveness of word $n$-grams decreases with longer units and larger $n$. These findings offer guidance for feature selection strategies and classification granularity in stylometric analysis.


\vspace{0.5em}
\noindent \textbf{Limitations and Future Directions.}  
Despite its effectiveness in English, our approach assumes structural and statistical regularities that may not generalize to languages with different typological properties. For example, languages with freer word order (e.g., Russian, Finnish), extensive inflection (e.g., Turkish, Latin), or logographic writing systems (e.g., Chinese) may not exhibit the same linear decay of sequential correlations or frequency distributions of stylistic features. This could affect the validity of the autocovariance model and the interpretability of $n$-gram-based features. Furthermore, our use of tf-idf and clustering assumes a stable tokenization scheme, which may break down in morphologically rich or script-divergent languages.

Additionally, while our use of the multivariate normal distribution offers computational tractability, it is a second-order approximation that may not capture higher-order sequential dependencies—such as long-range thematic recurrence or stylistic modulation across chapters. Extensions to more expressive models (e.g., non-Gaussian graphical models or sequence-aware deep embeddings) could improve sensitivity to such patterns.

Future work will focus on evaluating the cross-linguistic robustness of our framework and adapting it for morphologically complex or typologically diverse languages. We also aim to integrate language-specific preprocessing pipelines (e.g., lemmatization, morphological analysis) and explore alternative models for simulating correlated binary sequences with more flexible dependence structures.

In summary, our data-centric hypothesis-testing approach provides a principled method for quantifying the influence of sequential correlations in text classification. It contributes to a more refined understanding of how theme, style, and genre shape computational judgments and highlights the ongoing challenge of disentangling these components—especially in supervised settings. We believe this work lays the groundwork for more interpretable and language-adaptive tools in computational stylometry.

\section*{Acknowledgements} \label{acknowledgements} 

The research was funded in part by the Hebrew University of Jerusalem and 
by the European Union (ERC, MiDRASH, Project No.\@ 101071829). 
Views and opinions expressed are, however, those of the authors only and do not necessarily reflect those of the European Union or the European Research Council Executive Agency. 
Neither the European Union nor the granting authority can be held responsible for them.

We thank the anonymous referees for their valuable insight that helped improve this work.

\bibliographystyle{chicago}
\bibliography{Bibliography-MM-MC}

\clearpage

\appendix

\renewcommand{\thesection}{\Alph{section}}
\renewcommand{\thefigure}{\Alph{section}.\arabic{figure}}
\renewcommand{\thetable}{\Alph{section}.\arabic{table}}
\counterwithin{figure}{section}
\counterwithin{table}{section}

\section{Text Pairs Tables} \label{app_text_tabs}

\begin{table*}[pt!]
\scriptsize
    \centering
    \begin{tabular}{c c c c c} 
    \hline\hline
    Text Title & Author & Abbreviation \\
    \hline
    \textit{Harry Potter and the Philosopher's Stone} & J. K. Rowling & Potter\_1 \\
    \textit{Harry Potter and the Chamber of Secrets} & J. K. Rowling & Potter\_2 \\
    \textit{Harry Potter and the Prisoner of Azkaban} & J. K. Rowling & Potter\_3 \\
    \textit{Harry Potter and the Goblet of Fire} & J. K. Rowling & Potter\_4 \\
    \textit{Harry Potter and the Order of the Phoenix} & J. K. Rowling & Potter\_5 \\
    \textit{Harry Potter and the Half-Blood Prince} & J. K. Rowling & Potter\_6 \\
    \textit{Harry Potter and the Deathly Hallows} & J. K. Rowling & Potter\_7 \\
    \textit{Percy Jackson and the Olympians: The Lightning Thief} & Rick Riordan & Percy\_1 \\
    \textit{Percy Jackson and the Olympians: The Sea of Monsters} & Rick Riordan & Percy\_2 \\
    \textit{Percy Jackson and the Olympians: The Titan's Curse} & Rick Riordan & Percy\_3 \\
    \textit{Percy Jackson and the Olympians: The Demigod Files} & Rick Riordan & Percy\_5 \\
    \textit{The Lord of the Rings: The Fellowship of the Ring} & J. R. R. Tolkien & Lotr\_1 \\
    \textit{The Lord of the Rings: The Two Towers} & J. R. R. Tolkien & Lotr\_2 \\
    \textit{The Lord of the Rings: The Return of the King} & J. R. R. Tolkien & Lotr\_3 \\
    \textit{The Silmarillion} & J. R. R. Tolkien & Silmarillion \\
    \textit{David Copperfield} & Charles Dickens & Dickens\_Copperfield \\
    \textit{Oliver Twist} & Charles Dickens & Dickens\_OliverTwist \\
    \textit{Hard Time}s & Charles Dickens & Dickens\_HardTimes \\
    \textit{A Tale of Two Cities} & Charles Dickens & Dickens\_TwoCities \\
    \textit{The Pickwick Papers} & Charles Dickens & Dickens\_Pickwick \\
    \textit{The Adventures of Sherlock Holmes} & Sir Arthur Conan Doyle & Holmes\_1 \\
    \textit{The Memoirs of Sherlock Holmes} & Sir Arthur Conan Doyle & Holmes\_2 \\
    \textit{The Return of Sherlock Holmes} & Sir Arthur Conan Doyle & Holmes\_3 \\
    \textit{The Jungle Book} & Rudyard Kipling & Kipling\_JungleBook \\
    \textit{The Second Jungle Book} & Rudyard Kipling & Kipling\_JungleBook2 \\
    \textit{Barrack-Room Ballads} & Rudyard Kipling & Kipling\_Ballads \\
    \textit{The Chronicles of Narnia} & C. S. Lewis & Narnia \\
    \textit{Prince Caspian} & C. S. Lewis & Caspian \\
    \textit{The Screwtape Letters} & C. S. Lewis & Screwtape \\
    \textit{Emma} & Jane Austen & Austen\_Emma \\
    \textit{Pride and Prejudice} & Jane Austen & Austen\_PP \\
    \textit{Mansfield Park} & Jane Austen & Austen\_MansfieldPark \\
        Collection of works (spooky corpus) & Edgar Allen Poe & spooky\_EAP \\
        Collection of works (spooky corpus) & Marry Shelley & spooky\_MWS \\
        Collection of works (spooky corpus) & H. P. Lovecraft & spooky\_HPL \\
    \hline
    \end{tabular}
    \caption{All texts used in this work, their respective author, and abbreviation when referred to.}
    \label{app_tab_allTexts}
\end{table*}

\clearpage

\scriptsize
\begin{longtable}{|c|c|}
\hline
\textbf{Text (abbreviated)} & \textbf{Group Number} \\
\hline
\endfirsthead
\hline
\textbf{Text (abbreviated)} & \textbf{Group Number} \\
\hline
\endhead
\hline
\endfoot
\endlastfoot
Potter\_1 & 1 \\
Potter\_2 & 1 \\
Potter\_3 & 1 \\
Potter\_4 & 1 \\
Potter\_5 & 1 \\
Potter\_6 & 1 \\
Potter\_7 & 1 \\
Percy\_1 & 2 \\
Percy\_2 & 2 \\
Percy\_3 & 2 \\
Percy\_5 & 2 \\
Lotr\_1 & 3 \\
Lotr\_2 & 3 \\
Lotr\_3 & 3 \\
Silmarillion & 4 \\
Dickens\_Copperfield & 5 \\
Dickens\_OliverTwist & 5 \\
Dickens\_HardTimes & 5 \\
Dickens\_TwoCities & 5 \\
Dickens\_Pickwick & 5 \\
Holmes\_1 & 6 \\
Holmes\_2 & 6 \\
Holmes\_3 & 6 \\
Kipling\_JungleBook & 7 \\
Kipling\_JungleBook2 & 7 \\
Kipling\_Ballads & 8 \\
Narnia & 9 \\
Caspian & 9 \\
Screwtape & 10 \\
Austen\_Emma & 11 \\
Austen\_PP & 11 \\
Austen\_MansfieldPark & 11 \\
spooky\_EAP & 12 \\
spooky\_MWS & 13 \\
spooky\_HPL & 14 \\

\caption{Author-wise book groups from which multi-book samples of a single author are generated.}
\label{app_tab_textGroups}
\end{longtable}

\scriptsize
\begin{longtable}{|c|c|c|}

\hline
\textbf{Index} & \textbf{Text 1} & \textbf{Text 2} \\
\hline
\endfirsthead
\hline
\textbf{Index} & \textbf{Text 1} & \textbf{Text 2} \\
\hline
\endhead
\hline
\endfoot
\endlastfoot

1 & spooky\_HPL & Screwtape \\
2 & Lotr\_3, Lotr\_2, Lotr\_1 & Percy\_5 \\
3 & Austen\_Emma, Austen\_PP, Austen\_MansfieldPark & Screwtape \\
4 & Narnia, Caspian & spooky\_MWS \\
5 & Potter\_4, Potter\_2 & Silmarillion \\
6 & spooky\_MWS & spooky\_HPL \\
7 & spooky\_HPL & Potter\_7, Potter\_3, Potter\_2 \\
8 & spooky\_MWS & Lotr\_3 \\
9 & Dickens\_TwoCities, Dickens\_OliverTwist, Dickens\_Copperfield & Narnia, Caspian \\
10 & Austen\_Emma & Kipling\_JungleBook \\
11 & spooky\_MWS & Narnia, Caspian \\
12 & Kipling\_Ballads & Lotr\_1, Lotr\_2, Lotr\_3 \\
13 & Kipling\_Ballads & Percy\_1, Percy\_3 \\
14 & Potter\_4, Potter\_7, Potter\_2 & spooky\_HPL \\
15 & Silmarillion & Potter\_7 \\
16 & spooky\_MWS & spooky\_HPL \\
17 & spooky\_MWS & Kipling\_Ballads \\
18 & Lotr\_2 & Percy\_5 \\
19 & Narnia, Caspian & spooky\_HPL \\
20 & Potter\_6, Potter\_5, Potter\_3 & Holmes\_1 \\
21 & Austen\_Emma, Austen\_MansfieldPark, Austen\_PP & Dickens\_TwoCities, Dickens\_OliverTwist \\
22 & Percy\_2, Percy\_3, Percy\_1 & spooky\_HPL \\
23 & Potter\_1 & Kipling\_JungleBook2 \\
24 & Holmes\_1, Holmes\_3, Holmes\_2 & Silmarillion \\
25 & spooky\_EAP & Kipling\_Ballads \\
26 & spooky\_EAP & Potter\_3, Potter\_2 \\
27 & Narnia, Caspian & Kipling\_Ballads \\
28 & spooky\_HPL & spooky\_EAP \\
29 & Narnia, Caspian & Potter\_1, Potter\_3 \\
30 & Silmarillion & spooky\_HPL \\
31 & spooky\_HPL & Narnia, Caspian \\
32 & Kipling\_JungleBook & Screwtape \\
33 & Silmarillion & spooky\_MWS \\
34 & Lotr\_2, Lotr\_3 & Kipling\_JungleBook \\
35 & Lotr\_2 & Percy\_5, Percy\_2 \\
36 & Kipling\_Ballads & Silmarillion \\
37 & Holmes\_2, Holmes\_3 & spooky\_HPL \\
38 & Screwtape & Lotr\_3, Lotr\_1 \\
39 & spooky\_HPL & Percy\_3, Percy\_2, Percy\_5 \\
40 & Austen\_Emma, Austen\_MansfieldPark & spooky\_HPL \\
41 & Lotr\_1 & Austen\_PP, Austen\_MansfieldPark \\
42 & Dickens\_TwoCities & Lotr\_3, Lotr\_1, Lotr\_2 \\
43 & spooky\_HPL & spooky\_EAP \\
44 & Dickens\_TwoCities, Dickens\_HardTimes, Dickens\_Pickwick & Narnia, Caspian \\
45 & Screwtape & spooky\_MWS \\
46 & Narnia, Caspian & Silmarillion \\
47 & Screwtape & Narnia, Caspian \\
48 & Austen\_PP & Kipling\_JungleBook2, Kipling\_JungleBook \\
49 & Kipling\_Ballads & Narnia, Caspian \\
50 & Potter\_2 & Kipling\_JungleBook2 \\

\caption{Randomly chosen pairs of texts composed by different authors that are evaluated in this work with their respective indices.}
\label{app_tab_unsupervised_texts_differentAuthors}
\end{longtable}

\footnotesize
\begin{longtable}{|c|c|c|c|} 
\hline
\textbf{Index} & \textbf{Text 1} & \textbf{Text 2} & \textbf{Vary Genre-wise} \\
\hline
\endfirsthead

\hline
\textbf{Index} & \textbf{Text 1} & \textbf{Text 2} & \textbf{Vary Genre-wise} \\
\hline
\endhead

\hline
\endfoot

\endlastfoot

1  & Potter\_1               & Potter\_2               & No  \\
2  & Potter\_1               & Potter\_3               & No  \\
3  & Potter\_1               & Potter\_4               & No  \\
4  & Potter\_1               & Potter\_5               & No  \\
5  & Potter\_1               & Potter\_6               & No  \\
6  & Potter\_1               & Potter\_7               & No  \\
7  & Potter\_2               & Potter\_3               & No  \\
8  & Potter\_2               & Potter\_4               & No  \\
9  & Potter\_2               & Potter\_5               & No  \\
10 & Potter\_2               & Potter\_6               & No  \\
11 & Potter\_2               & Potter\_7               & No  \\
12 & Potter\_3               & Potter\_4               & No  \\
13 & Potter\_3               & Potter\_5               & No  \\
14 & Potter\_3               & Potter\_6               & No  \\
15 & Potter\_3               & Potter\_7               & No  \\
16 & Potter\_4               & Potter\_5               & No  \\
17 & Potter\_4               & Potter\_6               & No  \\
18 & Potter\_4               & Potter\_7               & No  \\
19 & Potter\_5               & Potter\_6               & No  \\
20 & Potter\_5               & Potter\_7               & No  \\
21 & Potter\_6               & Potter\_7               & No  \\
22 & Lotr\_1                 & Lotr\_2                 & No  \\
23 & Lotr\_1                 & Lotr\_3                 & No  \\
24 & Lotr\_1                 & Silmarillion            & Yes \\
25 & Lotr\_2                 & Lotr\_3                 & No  \\
26 & Lotr\_2                 & Silmarillion            & Yes \\
27 & Lotr\_3                 & Silmarillion            & Yes \\
28 & Percy\_1                & Percy\_2                & No  \\
29 & Percy\_1                & Percy\_3                & No  \\
30 & Percy\_1                & Percy\_5                & No  \\
31 & Percy\_2                & Percy\_3                & No  \\
32 & Percy\_2                & Percy\_5                & No  \\
33 & Percy\_3                & Percy\_5                & No  \\
34 & Dickens\_Copperfield    & Dickens\_OliverTwist    & ?   \\
35 & Dickens\_Copperfield    & Dickens\_HardTimes      & ?   \\
36 & Dickens\_Copperfield    & Dickens\_TwoCities      & ?   \\
37 & Dickens\_Copperfield    & Dickens\_Pickwick       & ?   \\
38 & Dickens\_OliverTwist    & Dickens\_HardTimes      & ?   \\
39 & Dickens\_OliverTwist    & Dickens\_TwoCities      & ?   \\
40 & Dickens\_OliverTwist    & Dickens\_Pickwick       & ?   \\
41 & Dickens\_HardTimes      & Dickens\_TwoCities      & ?   \\
42 & Dickens\_HardTimes      & Dickens\_Pickwick       & ?   \\
43 & Dickens\_TwoCities      & Dickens\_Pickwick       & ?   \\
44 & Holmes\_1               & Holmes\_2               & No  \\
45 & Holmes\_1               & Holmes\_3               & No  \\
46 & Holmes\_2               & Holmes\_3               & No  \\
47 & Kipling\_JungleBook     & Kipling\_JungleBook2    & No  \\
48 & Kipling\_JungleBook     & Kipling\_Ballads        & Yes \\
49 & Kipling\_JungleBook2    & Kipling\_Ballads        & Yes \\
50 & Narnia                  & Caspian                 & No  \\
51 & Narnia                  & Screwtape               & Yes \\
52 & Caspian                 & Screwtape               & Yes \\
53 & Austen\_Emma            & Austen\_PP              & ?   \\
54 & Austen\_Emma            & Austen\_MansfieldPark   & ?   \\
55 & Austen\_PP              & Austen\_MansfieldPark   & ?   \\

\caption{All permutations of same-author text pairs used in both experiments. Additionally, the fourth column lists our assumption regarding whether or not each pair of texts varies greatly in genre. When indecisive, we mark that pair with "?".}
\label{app_tab_unsupervised_texts}
\end{longtable}

\clearpage

\section{Classification Algorithms with our Hypothesis-Testing Approach} \label{app_algs}

\begin{algorithm}
\caption{\textbf{Experiment}: Unsupervised ($k$-means)} 
	\begin{algorithmic}[1]
            \State Initiate MCC score array $v$ of length 100.
            \State Initiate $p$-value array $u$ of length 100.
		\For {$i=1,\ldots,100$}
                \State Draw a subsample of $c \cdot \mathrm{min}\{m^{(\ell )}_1, m^{(\ell )}_2 \}$ text units ($T^{(\ell ,n,f)}_{\mathrm{subsample}}$).
			\State Apply $2$-means on $T^{(\ell ,n,f)}_{\mathrm{subsample}}$ and receive label sequence $L$.
                \State Compute MCC score between $L$ and $L^{\mathrm{(\mathrm{real})}}$ and store in $v$.
                \State Estimate $p$-value of the MCC score with hypothesis-testing routine (null distribution composed of 1000 simulations) and store in $u$.
		\EndFor
        \State Apply FDR correction to $u$ to receive an FDR-corrected array $u_{\mathrm{FDR}}$.
        \State Chosen MCC score and $p$-value for the combination of $n$ and $\ell$, respectively: $\bar{v}$, $\bar{u}_{\mathrm{FDR}}$.
	\end{algorithmic} 
\label{alg_unsup}
\end{algorithm}

\begin{algorithm}
	\caption{\textbf{Experiment}: Supervised (GI)} 
	\begin{algorithmic}[1]
            \State Initiate MCC score array $v$ of length 100.
            \State Initiate $p$-value array $u$ of length 100.
		\For {$i=1,\ldots,100$}
                \State Draw a subsample of $c \cdot \mathrm{min}\{m^{(\ell )}_1, m^{(\ell )}_2 \}$ text units for training ($T^{(\ell ,n,f)}_{\mathrm{train}}$), the rest are for testing ($T^{(\ell ,n,f)}_{\mathrm{test}}$).
			\State Compute two label-wise centroids of $T^{(\ell ,n,f)}_{\mathrm{train}}$ according to $L^{\mathrm{(\mathrm{real})}}$.
                \State Initiate label-association array $w \in \mathbb{R}^{100 \times |T^{(\ell ,n,f)}_{\mathrm{test}}|}$.
		      \For {$j=1,\ldots,100$}
                \State Draw a subsample of $f$, satisfying $f_{\mathrm{subsample}} \in f$ and $|f_{\mathrm{subsample}}| \approx 0.3 \cdot |f|$.
                \For {$k=1,\ldots,|T^{(\ell ,n,f)}_{\mathrm{test}}|$}
                \State Compute the $2$-norm of the difference between $T^{(\ell ,n,f_{\mathrm{subsample}})}_{\mathrm{test}, k}$ and each label-wise centroid, respectively.
                \State $w_{jk}$ = label of the centroid for which the $2$-norm was smaller.
                \EndFor
                \State Determine the label for every sample in $T^{(\ell ,n,f)}_{\mathrm{test}}$ based on the one with which it is most associated in $w$, and receive a label sequence $L$.
                \EndFor
                \State Compute MCC score between $L$ and $L^{\mathrm{(\mathrm{real})}}$ and store in $v$.
                \State Estimate $p$-value of the MCC score with hypothesis-testing routine and store in $u$.
		\EndFor
        \State Apply FDR correction $u$ to receive an FDR-corrected array $u_{\mathrm{FDR}}$
        \State Chosen MCC score and $p$-value for the combination of $n$ and $\ell$, respectively: $\bar{v}$, $\bar{u}_{\mathrm{FDR}}$
	\end{algorithmic} 
 \label{alg_sup}
\end{algorithm}

\begin{algorithm}
	\caption{\textbf{Experiment}: Supervised (Cosine Similarity)} 
	\begin{algorithmic}[1]
            \State Initiate MCC score array $v$ of length 100.
            \State Initiate $p$-value array $u$ of length 100.
		\For {$i=1,\ldots,100$}
                \State Draw a subsample of $c \cdot \mathrm{min}\{m^{(\ell )}_1, m^{(\ell )}_2 \}$ text units for training ($T^{(\ell ,n,f)}_{\mathrm{train}}$), the rest are for testing ($T^{(\ell,f)}_{\mathrm{test}}$).
			\State Compute two label-wise centroids of $T^{(\ell,f)}_{\mathrm{train}}$ according to $L^{\mathrm{(\mathrm{real})}}$.
                \State Initiate label-association array $w \in \mathbb{R}^{|T^{(\ell,f)}_{\mathrm{test}}|}$.
		      \For {$j=1,\ldots,|T^{(\ell,f)}_{\mathrm{test}}|$}
                \State $L_j = \mathrm{index}[\mathrm{max}(\mathrm{cos}(\mathrm{centroid}_1, T^{(\ell,f)}_{\mathrm{test},j}), \mathrm{cos}(\mathrm{centroid}_2, T^{(\ell,f)}_{\mathrm{test},j}))]$.
                \EndFor
                \State Compute MCC score between $L$ and $L^{\mathrm{(\mathrm{real})}}$ and store in $v$.
                \State Estimate $p$-value of the MCC score with hypothesis-testing routine and store in $u$.
		\EndFor
        \State Apply FDR correction $u$ to receive an FDR-corrected array $u_{\mathrm{FDR}}$
        \State Chosen MCC score and $p$-value for the combination of $n$ and $\ell$, respectively: $\bar{v}$, $\bar{u}_{\mathrm{FDR}}$
	\end{algorithmic} 
 \label{alg_cosine}
\end{algorithm}

\section{Z-Score Embedding Based on Burrows’ Delta} \label{app_zscore}

To complement traditional tf-idf embeddings, we implement a stylometric embedding inspired by Burrows’ Delta \citep{burrows2002delta}, reformulated for compatibility with clustering algorithms operating in Euclidean space.

Let $N$ be the number of text units in a given comparison (e.g., blocks of $\ell$ words), and let $f$ denote the number of most frequent $n$-grams selected across the corpus. We construct a document-feature matrix $X \in \mathbb{R}^{N \times f}$, where each entry $x_{ij}$ corresponds to the raw frequency of the $j$th $n$-gram in the $i$th text unit.

We then standardize $X$ column-wise to obtain the Z-score matrix $Z \in \mathbb{R}^{N \times f}$, whose entries are given by:

\[
Z_{ij} = \frac{x_{ij} - \mu_j}{\sigma_j}
\]
where $\mu_j = \frac{1}{N} \sum_{i=1}^N x_{ij}$ is the mean frequency of feature $j$, and $\sigma_j$ is the corresponding standard deviation. In cases where $\sigma_j = 0$, the column is treated as constant and replaced with zeros in $Z$ to ensure numerical stability.

This transformation yields a stylometric embedding in which each feature reflects how unusually frequent a given $n$-gram is relative to its corpus-wide baseline. 
We apply $k$-means clustering directly to the matrix $Z$, using Euclidean distance in the standardized feature space. The resulting cluster assignments are then evaluated against ground truth labels using the Matthews correlation coefficient (MCC), and their statistical significance is assessed via the hypothesis-testing framework described in Section~\ref{hp_methodology}.

\section{Traditional Embeddings Significance Maps} \label{app_trad}

\subsection{Unsupervised Classification Results} \label{app_trad_us}


\begin{figure}[t!]
\centering
\rotatebox[origin=c]{90}{\includegraphics[scale = 0.105]{chars_diff_kmeans_batch-0.2.pdf}}
  \caption{Significance map for the attempt to apply $2$-means classification to distinguish 50 pairs of texts composed by different authors (see Table \ref{app_tab_unsupervised_texts_differentAuthors}), embedded using character $k$-mers with $f$ = 300, similarly to Figure \ref{Fig_res_words}. }
\end{figure}

\begin{figure}[t!]
\centering
\rotatebox[origin=c]{90}{\includegraphics[scale = 0.105]{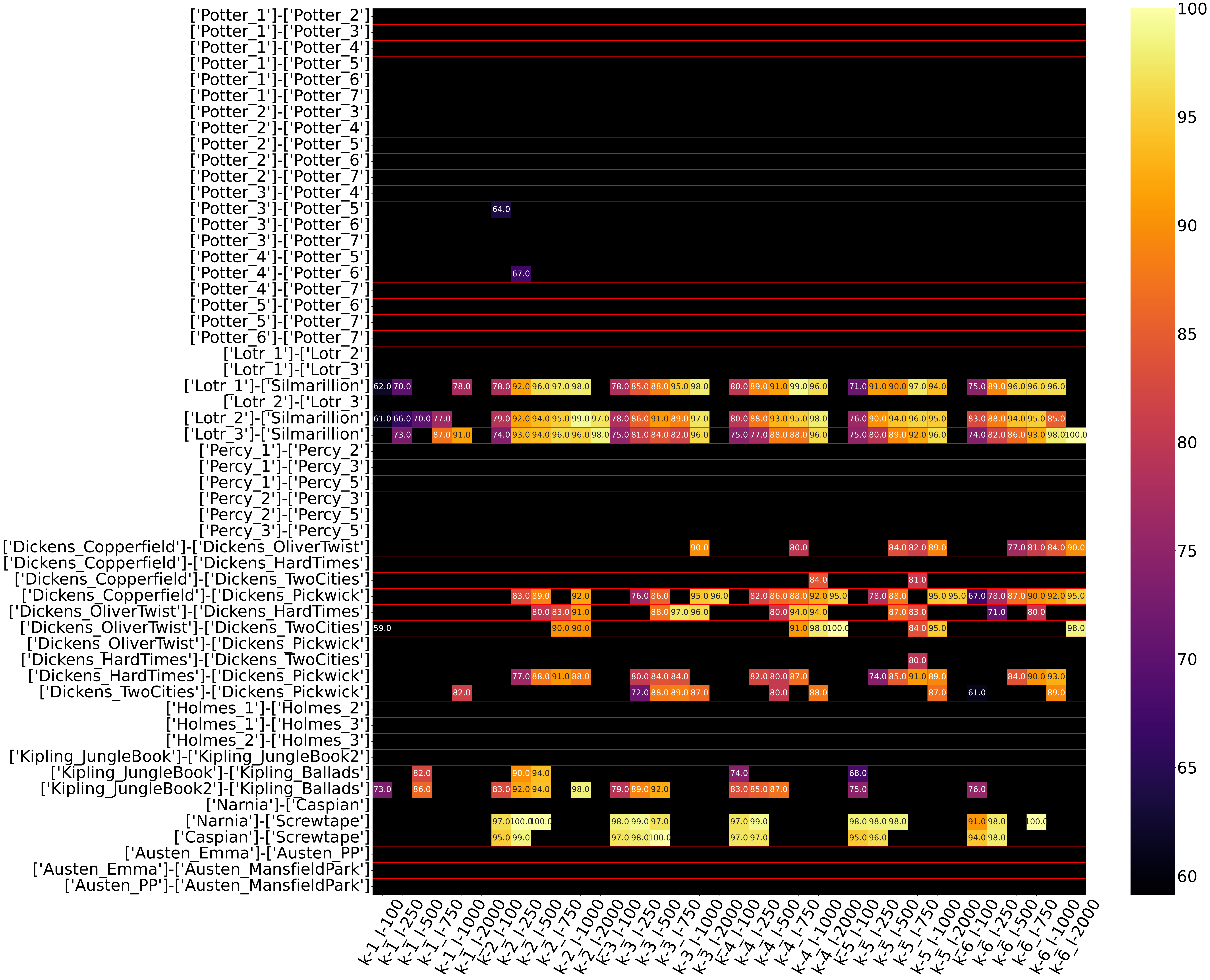}}
  \caption{Significance map for the attempt to apply $2$-means classification to distinguish all permutations of pairs of texts composed by the same author (see Table \ref{app_tab_unsupervised_texts}), embedded using character $k$-mers with $f$ = 300, similarly to Figure \ref{Fig_res_words}.}
\end{figure}

\begin{figure}[t!]
\centering
\rotatebox[origin=c]{90}{\includegraphics[scale = 0.105]{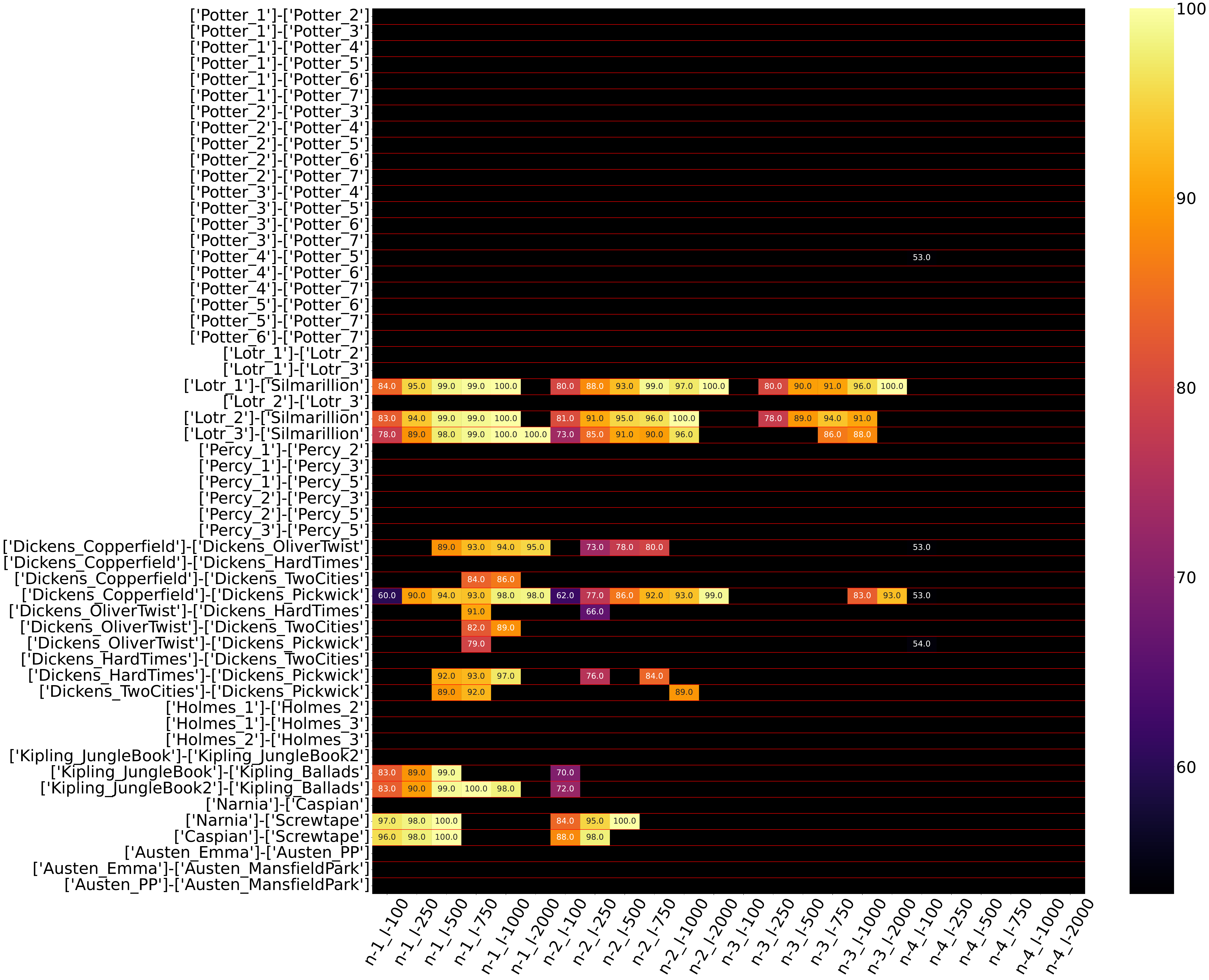}}
  \caption{Significance map for the attempt to apply $2$-means classification to distinguish all permutations of pairs of texts composed by the same author (see Table \ref{app_tab_unsupervised_texts}), embedded using word $n$-grams with $f$ = 300, similarly to Figure \ref{Fig_res_words}.}
\end{figure}

\begin{figure}[t!]
\centering
\rotatebox[origin=c]{90}{\includegraphics[scale = 0.105]{delta_diff_ngrams.pdf}}
  \caption{Significance map for the attempt to apply $2$-means classification to distinguish 50 pairs of texts composed by different authors (see Table \ref{app_tab_unsupervised_texts_differentAuthors}), embedded using word $n$-grams and Z-score embedding with $f$ = 300, similarly to Figure \ref{Fig_res_words}. }
     \label{Fig_delta_diff_ngrams}
\end{figure}

\begin{figure}[t!]
\centering
\rotatebox[origin=c]{90}{\includegraphics[scale = 0.105]{delta_diff_kmers.pdf}}
  \caption{Significance map for the attempt to apply $2$-means classification to distinguish 50 pairs of texts composed by different authors (see Table \ref{app_tab_unsupervised_texts_differentAuthors}), embedded using character $k$-mers and Z-score embedding with $f$ = 300, similarly to Figure \ref{Fig_res_words}. }
     \label{Fig_delta_diff_kmers}
\end{figure}

\begin{figure}[t!]
\centering
\rotatebox[origin=c]{90}{\includegraphics[scale = 0.105]{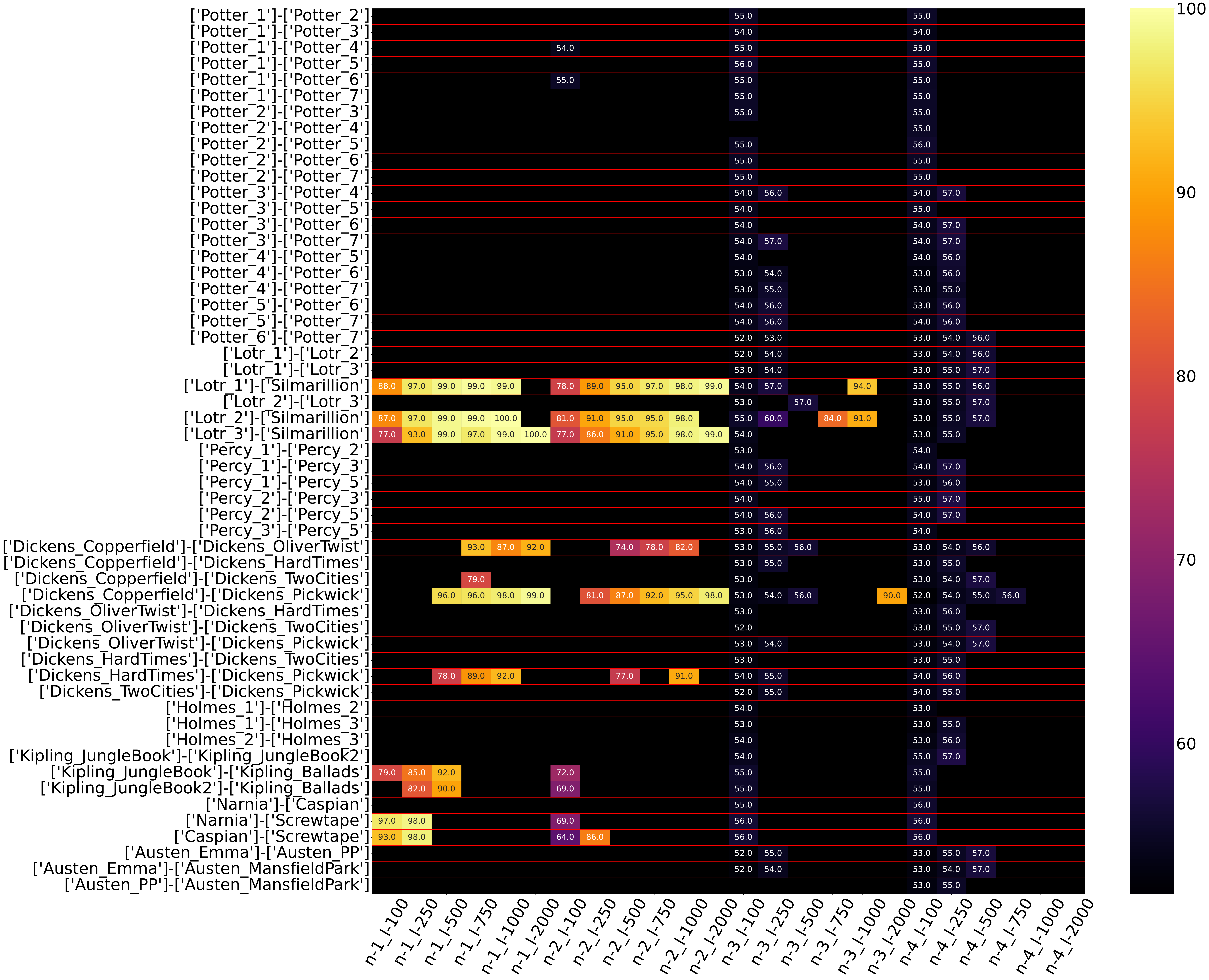}}
  \caption{Significance map for the attempt to apply $2$-means classification to distinguish all permutations of pairs of texts composed by the same author (see Table \ref{app_tab_unsupervised_texts}), embedded using word $n$-grams and Z-score embedding with $f$ = 300, similarly to Figure \ref{Fig_res_words}. }
     \label{Fig_delta_same_ngrams}
\end{figure}

\begin{figure}[t!]
\centering
\rotatebox[origin=c]{90}{\includegraphics[scale = 0.105]{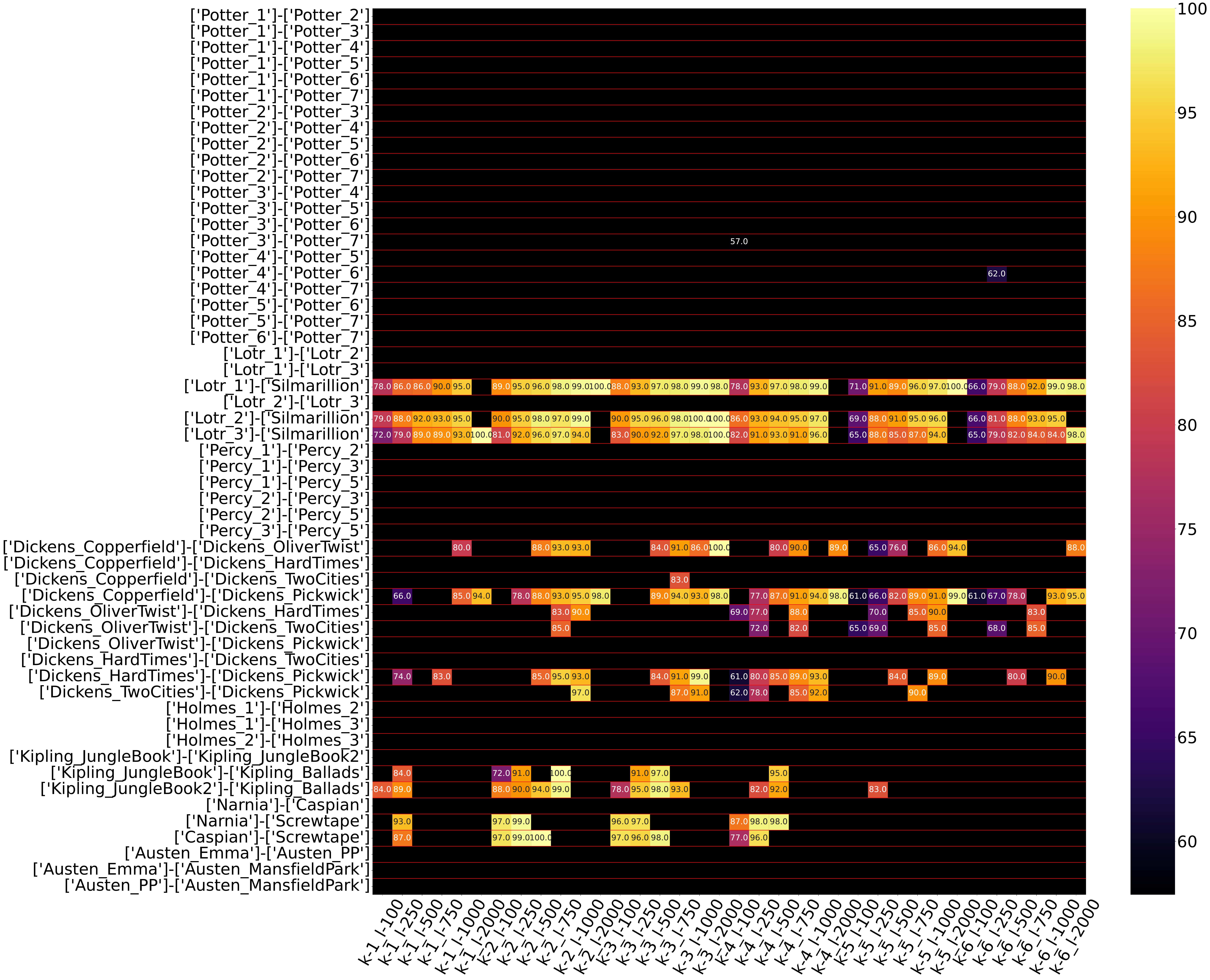}}
  \caption{Significance map for the attempt to apply $2$-means classification to distinguish all permutations of pairs of texts composed by the same author (see Table \ref{app_tab_unsupervised_texts}), embedded using character $k$-mers and Z-score embedding with $f$ = 300, similarly to Figure \ref{Fig_res_words}. }
     \label{Fig_delta_same_kmers}
\end{figure}

\clearpage
\subsection{Supervised Classification Results} \label{app_trad_s}

\begin{figure}
\centering
\rotatebox[origin=c]{90}{\includegraphics[scale = 0.105]{words_diff_GI_batch-0.5.pdf}}
  \caption{Significance map for the attempt to apply GI supervised classification to distinguish 50 pairs of texts composed by different authors (see Table \ref{app_tab_unsupervised_texts_differentAuthors}), embedded using word $n$-grams with $f$ = 300, similarly to Figure \ref{Fig_res_words}. }
\end{figure}

\begin{figure}
\centering
\rotatebox[origin=c]{90}{\includegraphics[scale = 0.105]{chars_diff_GI_batch-0.5.pdf}}
  \caption{Significance map for the attempt to apply GI supervised classification to distinguish 50 pairs of texts composed by different authors (see Table \ref{app_tab_unsupervised_texts_differentAuthors}), embedded using character $k$-mers with $f$ = 300, similarly to Figure \ref{Fig_res_words}. }
\end{figure}

\begin{figure}
\centering
\rotatebox[origin=c]{90}{\includegraphics[scale = 0.105]{words_same_GI_batch-0.5.pdf}}
  \caption{Significance map for the attempt to apply GI supervised classification to distinguish all permutations of pairs of texts composed by the same author (see Table \ref{app_tab_unsupervised_texts}), embedded using word $n$-grams with $f$ = 300, similarly to Figure \ref{Fig_res_words}.}
\end{figure}

\begin{figure}
\centering
\rotatebox[origin=c]{90}{\includegraphics[scale = 0.105]{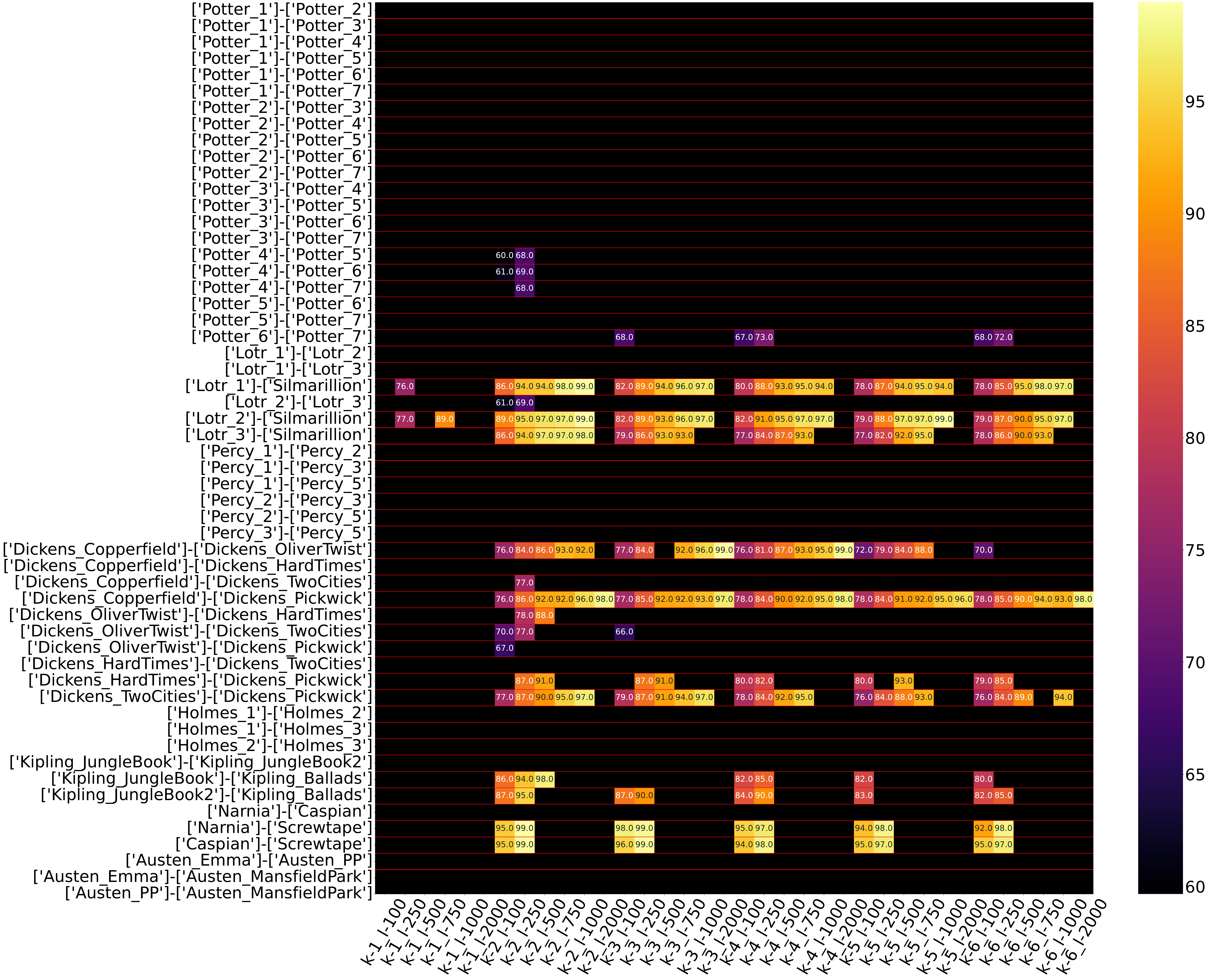}}
  \caption{Significance map for the attempt to apply GI supervised classification to distinguish all permutations of pairs of texts composed by the same author (see Table \ref{app_tab_unsupervised_texts}), embedded using character $k$-mers with $f$ = 300, similarly to Figure \ref{Fig_res_words}.}
\end{figure}

\clearpage
\section{Neural Embeddings Significance Maps} \label{app_neural}

\subsection{Unsupervised Classification Results} \label{app_neural_us}

\begin{figure}
\centering
\rotatebox[origin=c]{90}{\includegraphics[scale = 0.105]{neural_diff_kmeans_batch-0.2.pdf}}
  \caption{Significance map for the attempt to apply $2$-means classification to distinguish 50 pairs of texts composed by different authors (see Table \ref{app_tab_unsupervised_texts_differentAuthors}), embedded using STAR, similarly to Figure \ref{Fig_res_words}.}
\end{figure}

\begin{figure}
\centering
\rotatebox[origin=c]{90}{\includegraphics[scale = 0.105]{neural_same_kmeans_batch-0.2.pdf}}
  \caption{Significance map for the attempt to apply $2$-means classification to distinguish all permutations of pairs of texts composed by the same author (see Table \ref{app_tab_unsupervised_texts}), embedded using STAR, similarly to Figure \ref{Fig_res_words}.}
\end{figure}

\clearpage
\subsection{Supervised Classification Results} \label{app_neural_s}

\begin{figure}
\centering
\rotatebox[origin=c]{90}{\includegraphics[scale = 0.105]{neural_diff_cosine_batch-0.5.pdf}}
  \caption{Significance map for the attempt to apply supervised cosine similarity classification to distinguish 50 pairs of texts composed by different authors (see Table \ref{app_tab_unsupervised_texts_differentAuthors}), embedded using STAR, similarly to Figure \ref{Fig_res_words}.}
\end{figure}

\begin{figure}
\centering
\rotatebox[origin=c]{90}{\includegraphics[scale = 0.105]{neural_same_kmeans_batch-0.2.pdf}}
  \caption{Significance map for the attempt to apply supervised cosine similarity classification to distinguish all permutations of pairs of texts composed by the same author (see Table \ref{app_tab_unsupervised_texts}), embedded using STAR, similarly to Figure \ref{Fig_res_words}.}
          \label{Fig_res_words_end}
\end{figure}

\end{document}